\documentclass[letterpaper,twocolumn,10pt]{article}
\usepackage{usenix-2020-09}

\usepackage{times}
\usepackage{soul}
\usepackage{url}

\usepackage[utf8]{inputenc}
\usepackage[small]{caption}
\usepackage{graphicx}
\usepackage{amsmath}
\usepackage{amsthm}
\usepackage{booktabs}
\usepackage{todonotes}
\usepackage{hhline}
\usepackage{multirow}
\usepackage{colortbl}
\usepackage{amsfonts}       
\usepackage{cleveref}
\usepackage{subfig}

\usepackage{xcolor}
\usepackage[export]{adjustbox}
\usepackage[ruled,vlined]{algorithm2e}
\usepackage[noend]{algpseudocode}

\usepackage{xspace}

\newcommand{\ie}{\textit{i.e.,}\@\xspace}
\newcommand{\eg}{\textit{e.g.,}\@\xspace}
\newcommand{\etal}{\textit{et al.}\@\xspace}
\newcommand{\realnum}{{\rm I\!R}}
\newcommand\norm[1]{\left\lVert#1\right\rVert}

\newcommand{\baiwu}[1]{\textcolor{blue}{Baiwu: #1}}

\newcommand{\others}[1]{\textcolor{cyan}{others: #1}}

\begin{document}

\date{}

\title{On Attribution of Deepfakes}

\author{
{\rm Baiwu Zhang}\\
University of Toronto and Vector Institute
\and
{\rm Jin Peng Zhou}\\
University of Toronto and Vector Institute
\and
{\rm Ilia Shumailov}\\
University of Toronto, Vector Institute and University of Cambridge
\and
{\rm Nicolas Papernot}\\
University of Toronto and Vector Institute
} 

\maketitle

\begin{abstract}
Progress in generative modelling, especially generative adversarial networks, have made it possible to efficiently synthesize and alter media at scale. Malicious individuals now rely on these machine-generated media, or \textit{deepfakes}, to manipulate social discourse. In order to ensure media authenticity, existing research is focused on deepfake detection. Yet, the adversarial nature of frameworks used for generative modeling suggests that progress towards detecting deepfakes will enable more realistic deepfake generation. Therefore, it  comes at no surprise that developers of generative models are under the scrutiny of stakeholders dealing with misinformation campaigns. At the same time, generative models have a lot of positive applications. As such, there is a clear need to develop tools that ensure the transparent use of generative modeling, while minimizing the harm caused by malicious applications. 
    
Our technique optimizes over the source of entropy of each generative model to probabilistically \textit{attribute} a deepfake to one of the models. We evaluate our method on the seminal example of face synthesis, demonstrating that our approach achieves 97.62\% attribution accuracy, and is less sensitive to perturbations and adversarial examples. We discuss the ethical implications of our work, identify where our technique can be used, and highlight that a more meaningful legislative framework is required for a more transparent and ethical use of generative modeling. Finally, we argue that model developers should be capable of claiming plausible deniability and propose a second framework to do so -- this allows a model developer to produce evidence that they did not produce media that they are being accused of having produced.

\end{abstract}
\section{Introduction}



For centuries, humans have forged content. In 1777, counterfeit letters were crafted to make it look like General Washington did not want 
to fight against Great Britain~\cite{george_washington}. Computer software such as Adobe Photoshop has long made it possible to edit photos in realistic ways. The advent of artificial intelligence and machine learning (ML) has exacerbated the issue. For example, recent progress in \textit{generative modeling} pushes feasibility of forgery into the realm of video, while also enabling sophisticated image forgery at scale. Some negative applications of generative modeling have been discussed in the media: revenge pornography~\cite{wang_2019}, misinformation campaigns by politicians~\cite{jee_2020}, fake journalism~\cite{vincent_2020,futurism_2020}, and impersonating company executives for profit~\cite{stupp_2019}. Beyond individuals that are directly targeted by the forged content, societal consequences include a growing lack of trust in content shared electronically---affecting all forms of human communication. 


It is attractive to develop detection schemes capable of automatically identifying and flagging content that was machine-manipulated; we collectively call such content \textit{deepfakes} through the remainder of the paper. This would enable social media platforms like Facebook to remove deepfakes without having to manually check each suspected post~\cite{facebook_2020}. One may even consider applying ML to the very problem of detecting deepfakes. However, this is unlikely to lead to a solution capable of robustly identifying them: given access to tools for detecting such content, the algorithms used to create deepfakes can learn to produce even better content. For example, generative adversarial networks (GANs), a common type of generative models, are designed around the core idea of training a generative model by optimizing an objective which rewards defeating a detector (or discriminator) of fake content. Additionally, ML-based deepfake detectors are vulnerable to adversarial examples~\cite{carlini2020evading} which can be found using generic approaches~\cite{biggio2013evasionattacks,szegedy2013intriguing}. Even in the face of progress in generative modeling, manual forensics is still capable of distinguishing content manipulated by a machine. Yet, automating this process is unlikely to yield robust results.

To make progress on this front, we tackle this pressing social issue through a different lens and seek to attribute generated images to generative models. That is, we design mechanisms that, given an image, will retrieve the original model that generated that image.
Our method characterizes the mapping learned by each generator between its source of entropy and the media that it produces. Given an image and a set of candidate generators, we formulate an optimization problem to attempt to reconstruct a random seed which could have led each generator to produce the image. Our analysis suggests that, with high probability, a seed is best reconstructed on the generator which originally produced the image, even if the recovered seed is different from the seed originally used. This yields a signature which forensics experts can use to aid their investigations to attribute the origin of deepfakes. 


Next, we study the robustness of this attribution mechanism in the face of adversaries. We find that deepfakes can be manipulated post-generation to increase the difficulty of reconstructing the corresponding seed. This limitation makes it difficult for our approach to attribute deepfakes to specific generators with strict guarantees of integrity. Our approach is nevertheless less sensitive to perturbations crafted by an adversary than prior work on detecting deepfakes~\cite{frank2020leveraging}. However, we show that certain GANs can generate any image as long as they are provided with sufficiently high-entropy inputs. This, in turn, means that when precision is critical, an attribution mechanism cannot be relied upon to provide plausible deniability. We discuss the limitations of our method and highlight that additional mechanisms are required to support attribution with strict guarantees. We provide an example of such mechanism in the Appendix.



To summarize, our contributions are:

\begin{enumerate}
\itemsep0em
    \item We define attribution of synthetic images to  generative models and reason about its theoretical and practical feasibility  (\S~\ref{sec:pb-def}). 
    \item We discuss ethical implications of GAN attribution techniques (\S~\ref{sec:ethics}) and develop a realistic threat model for GAN attribution (\S~\ref{sec:threat_model}).
    \item We propose (\S~\ref{sec:relaxed_integrity}) and evaluate (\S~\ref{sec:eval}) a seed reconstruction approach to assist forensics in the attribution of synthesized samples to a specific GAN, when relaxed integrity is permissible. 
    \item Our method can be used to identify the GAN that produced the deepfake in more than 90\% of cases, even distinguishing GANs that are just a few gradient steps apart. Through a user study, we confirm that human participants agree with the results of attribution 93.7\% of the time, suggesting that the output of our approach can be interpreted by a human. 
    \item To show our approach cannot be solely relied upon to provide plausible deniability to model developers, we study its robustness to post-generation manipulations of deepfakes. We experiment with \textit{unbounded} perturbations, both non-adversarial (\S~\ref{sec:no_adv_manip}) and adversarial (\S~\ref{sec:adv_manip}).
    \item We describe limitations of our method and describe additional mechanisms required for a model developer to obtain plausible deniability in a strict integrity setting. 
   
\end{enumerate}

Before we dive in the technical details, we first discuss the ethical implications of our research in the following section; this area raises a number of important societal questions.

\section{Ethical considerations} 
\label{sec:ethics}

While our paper introduces techniques designed to address some of the issues raised by deepfakes, it is important to proactively consider how these techniques will be used in practice as well as understand what their limitations are.

\paragraph*{Correctness} As highlighted in our work on StyleGAN and StyleGAN2 in \S~\ref{sec:problem_def:limits}, some machine-manipulated content produced by generative adversarial networks (GANs) are difficult to attribute. This may occur when different generators are trained with similar datasets and architectures for instance. To avoid a false sense of security, we take care in our evaluation to consider both false positive (FP) and false negative (FN) rates when attributing a deepfake to a specific generator. Different stakeholders will pay more attention to the former (\eg social media platform) or the latter (\eg a court analyzing evidence).  For this reason, we argue in \S~\Cref{sec:strict_integrity} that in some cases model developers will have to resort to more traditional operational security to achieve plausible deniability. 

\paragraph*{Unintended uses} While this is not the intended use case, our technology could be used by state actors to justify censorship of legitimate content shared by humans. We hope that by bringing this issue to the attention of our community, potential victims will be better informed of the risks involved when using generative ML to evade censorship.

\paragraph*{Bias} As our approach relies on the underlying GANs for attribution, it may underperform when some of the GANs themselves are unable to generate media from certain populations because of limitations with their training algorithm and/or data. This may lead to an unintended lack of fairness in attribution results of our seed reconstruction approach. 

\section{Problem definition}
\label{sec:pb-def}

Generative models differ from the more commonly known ML classifiers in a number of ways. Typically, a classifier is trained in a \textit{supervised} manner to map semantically meaningful samples from a high-dimensional space (\eg images) to a small number of classes (\eg a label corresponding to a specific object in an object recognition task). Instead, generative models are trained in an \textit{unsupervised} manner and map a low-dimensional random vector called the \textit{seed} to a semantically meaningful sample that belongs to a high-dimensional space. More precisely, generative models are trained to model a data distribution $p_{data}$ as follows: a prior $p_{\bf{s}}$ is defined for the random distribution of seeds. The generative model then maps seeds sampled from this prior ${\bf{s}} \sim p_{\bf{s}}$  to the space of semantically meaningful samples by computing the output of a generative model $G(\bf{s})$. The goal of generative modeling is to train a model $G$, which when fed with different seeds ${\bf{s}}\sim p_{\bf{s}}$, synthesizes samples $\mathbf{x}$ such that it may seem like $\mathbf{x} \sim p_{data}$.

Like the rest of ML, generative modeling has benefited from advances in deep learning. While a survey of generative modeling is outside the scope of this work, we refer readers to the tutorial by  Goodfellow~\cite{goodfellow2016nips}. Progress in generative modeling has opened the door for exciting applications. For instance, a piece of a machine-generated painting was sold for \$430,000 at an auction~\cite{ajdellinger_2019}, text and synthesized photos can now be bidirectionally translated~\cite{zhang2017stackgan}, and 3D models can be generated from a single 2D image~\cite{chen2019learning} to name a few. Two prominent techniques for generative modeling include variational autoencoders (VAEs)~\cite{DBLP:journals/corr/abs-1906-00446,pu2016variational,vahdat2020nvae} and generative adversarial networks (GANs)~\cite{goodfellow2016nips,NIPS2016_6399,8253599,arjovsky2017principled,dieng2019prescribed,radford2015unsupervised}. We focus on the latter approach since (i) GANs have seen widespread utilization in generation of (human-perceived) high-quality and high-resolution synthetic media such as images~\cite{DBLP:journals/corr/abs-1904-01121}, videos~\cite{tulyakov2018mocogan}, and audio~\cite{donahue2018adversarial}, and until very recently~\cite{vahdat2020nvae}, VAEs were unable to match the quality of media produced by GANs; (ii) despite their numerous useful applications, GANs are also the basis for most of the approaches behind deepfakes~\cite{frank2020leveraging,tolosana2020deepfakes} owing to their ease of adaptation 
to variants of pure image synthesis that are commonly employed in deepfake creation like inpainting~\cite{yeh2017semanticinpainting}, style transfer~\cite{huang2017arbitrarystyletransfer}, or face swapping~\cite{nirkin2019fsgan}. 





\subsection{Primer on GANs}
\label{ssec:gan-primer}




Generative Adversarial Networks (GANs) consist of a generator $\mathcal{G}$ and a discriminator $\mathcal{D}$. The generator, parameterized by $\vartheta^{(\mathcal{G})}$, takes 
a latent variable (\ie seed) ${\bf{s}}\in\realnum^{d}$ as input, 
and outputs an observed variable ${\mathbf{x}}\in\realnum^{m}$\footnote{$\realnum^m$ is also referred to as $\mathbf{X}$, the space of images}, where ${\mathbf{x}}=\mathcal{G}({\bf{s}})$. The discriminator, parametrized by $\vartheta^{(\mathcal{D})}$, 
takes an observed variable ${\mathbf{x}}\in\realnum^{m}$ as input, 
and outputs a score $\mathcal{D}({\mathbf{x}})\in\realnum$ that quantifies the probability that ${\mathbf{x}}$ was synthesized by $\mathcal{G}$. Both $\mathcal{G}$ and $\mathcal{D}$ have a cost function $J_{\mathcal{G}}$ and $J_{\mathcal{D}}$ respectively that they aim to minimize:
\begin{equation}
    J_{\mathcal{D}} = \mathbb{E}_{\mathbf{x} \sim p_{data}}[-log \mathcal{D}(\mathbf{x})] + \mathbb{E}_{\mathbf{s} \sim p_{\bf{s}}}[-log (1 - \mathcal{D}(\mathcal{G}(\mathbf{s})))]
\end{equation}
\begin{equation}
    J_{\mathcal{G}} = \mathbb{E}_{\mathbf{s} \sim p_{\bf{s}}}[log (1 - \mathcal{D}(\mathcal{G}(\mathbf{s})))]
\end{equation}
Although both $J_{\mathcal{D}}$ and $J_{\mathcal{G}}$ depend on $\vartheta^{(\mathcal{G})}$ and $\vartheta^{(\mathcal{D})}$, the generator and discriminator can only control their own parameters while minimizing their respective cost functions.



In this paper, 
we are primarily interested in GANs for 
image generation. Specifically, the generator outputs synthetic RGB images given latent variables. The discriminator takes
either real or synthetic images 
as input, 
and outputs a score that classifies them as either real or synthetic. 
During 
the training process,
the generator tries to trick the discriminator by producing seemingly realistic images, whereas the discriminator aims to distinguish between real and synthetic images to counter the generator. Note that the GANs we study here are already trained to convergence and we simply use them to generate synthetic samples without performing any update to either $\vartheta^{(G)}$ or $\vartheta^{(D)}$.

\subsection{Attributing deepfakes to GANs}
\label{sec:problem_def}


Recall that in the introduction we outlined how the problem of detecting content that was machine-manipulated is ill-defined and unlikely to yield advances towards mitigating deepfakes in practice, because better detection is likely to spur progress in generation. Indeed, we saw in \S~\ref{ssec:gan-primer} how the GAN framework itself integrates a detector for fake content---the discriminator. Instead, we assume that the sample ${\mathbf{x}}$ is already known to be fake and focus on the problem of attribution. 

We define \textit{\textbf{attribution}} of a sample to a generator as the post hoc association of this sample to the generator model that originally generated it. Using the notation above, if ${\mathbf{x}}=g({\bf{s}})$ for a seed ${\bf{s}}\sim p_{\bf{s}}$, then ${\mathbf{x}}$ is said to be \textbf{\textit{attributed}} to $g$. Given an artificial sample ${\bf{s}}$ and a set of generative models $ G = \{g_{0}, g_{1}, ..., g_{n}\}$, we say that the \textit{attribution} was successful if the sample is attributed to the original model and not any other model. Here we assume that the model which generated the sample ${\mathbf{x}}$ is within the set $G$. In practice, our seed reconstruction approach from \S~\ref{sec:relaxed_integrity} ranks different candidate models, so we consider the attribution successful when the model in $G$ which generated the sample ${\mathbf{x}}$ is ranked first. Note how the problem of attribution can be recast as a classification task under the ML terminology (see for instance~\cite{frank2020leveraging}), where each class corresponds to the index $i\in \{1..n\}$ of the generator model $g_i\in G$. We use this analogy later to evaluate the robustness of our attribution mechanism to deepfakes specifically crafted by adversaries to mislead attribution. 

Our proposed definition of attribution goes back to the \textit{authenticity} property in classical computer security literature; we are authenticating data using knowledge of the  generator that produced it. Authenticity is defined as a combination of three properties: integrity; freshness; and authorship by a right principal~\cite{anderson2008security}. Whilst this paper largely considers the question of attribution, we must understand integrity, and the implications it has on the broader problem of plausible deniability when dealing with deepfakes. Problem of attribution can be split into two subproblems: 

\begin{enumerate}
    \item \textit{Strict attribution:} When integrity is strictly guaranteed, only a sample ${\mathbf{x}}$ (\eg image) which was produced by a specific GAN $g$ should be attributed to $g$. The process needs to favor \textit{precision}: the smallest perturbation to a sample and/or model parameters should lead to the attribution process failing for ${\mathbf{x}}$ and $g$.
    
    \item \textit{Relaxed attribution:} On the other hand, integrity can be enforced in a relaxed manner, akin to malleability, where one wants attribution to model $g$ to succeed even if the sample ${\mathbf{x}}$ was perturbed after it was generated by model $g$. This covers cases where \textit{recall} is of importance---small changes should not lead to an incorrect attribution. In this setting, it becomes essential to define what small means and limit the perturbation for which the attribution process is expected to be tolerant, \eg within an $l_{\infty}$-ball. 
\end{enumerate}
 
Both scenarios described above have real-life applications. The former considers liability cases, where artificial content should only be attributed to the generator that produced them. For example, imagine a GAN designed to generate videos of humans that was fine-tuned by an offender to generate child abuse materials. In such a scenario, creators of the human-generating GANs should not be held liable for the fine-tuned GAN, but rather the offender should be prosecuted. The latter case considers digital forensics and digital right management infringement cases. For example, one might want to prove that GAN-produced content is copyrighted and transformations applied to the content (as in research into watermarking, fingerprinting, and general information hiding~\cite{petitcolas1999infohidingsurvey}) should not disrupt attribution. Transformations can range from being adversarial to random, and include different encoding schemes. 


In this paper we focus on providing a solution to relaxed attribution. Additionally, we recognize that strict attribution is not always possible in~\S~\Cref{sec:problem_def:limits}, and its precise nature makes machine learning an unsuitable tool. Furthermore, we discuss plausible deniability as one objective of strict attribution in~\S~\Cref{sec:problem_def:pd}, which could be achieved with a cryptographic solution as we briefly touch upon in~\S~\Cref{sec:strict_integrity} and leave for future work.

\subsection{Why is it possible to attribute synthesized samples to GANs?}


It is not immediately clear that attribution should be possible in the first place; if multiple GANs were able to converge perfectly (\ie model the data distribution $p_{data}$ exactly), it should be theoretically impossible to attribute a sample ${\mathbf{x}}$ to one of these GANs. 
Indeed, GANs learn to approximate a data distribution $p_{data}$ by continuously interacting with a discriminator whose role is to measure how far synthetic samples, generated by the generator, are from the data distribution $p_{data}$. 
In the limit of unlimited samples from $p_{data}$ and unbounded model capacity (\ie enough neurons to approximate any function), 
different GANs solving the same task should learn to produce indistinguishable synthetic data 
according to some distribution $p_{\mathcal{G}}$, such that $p_{\mathcal{G}} = p_{data}$.
Yet, in reality the optimization problem being solved exhibits multiple optima---datasets are noisy and only partially describe the data distribution $p_{data}$ at hand, whereas imperfect generator architectures end up describing data manifolds only to a certain extent. This makes it difficult for GANs to converge exactly to the underlying data distribution, as the training procedure approximates the optimization problem and is susceptible to artifacts of learning such as the initialization of the model's parameters at the beginning of training for instance. All together this leads to GANs that are intrinsically \textit{biased}.


\begin{figure}[t]
  \centering
  \includegraphics[width=0.70\linewidth]{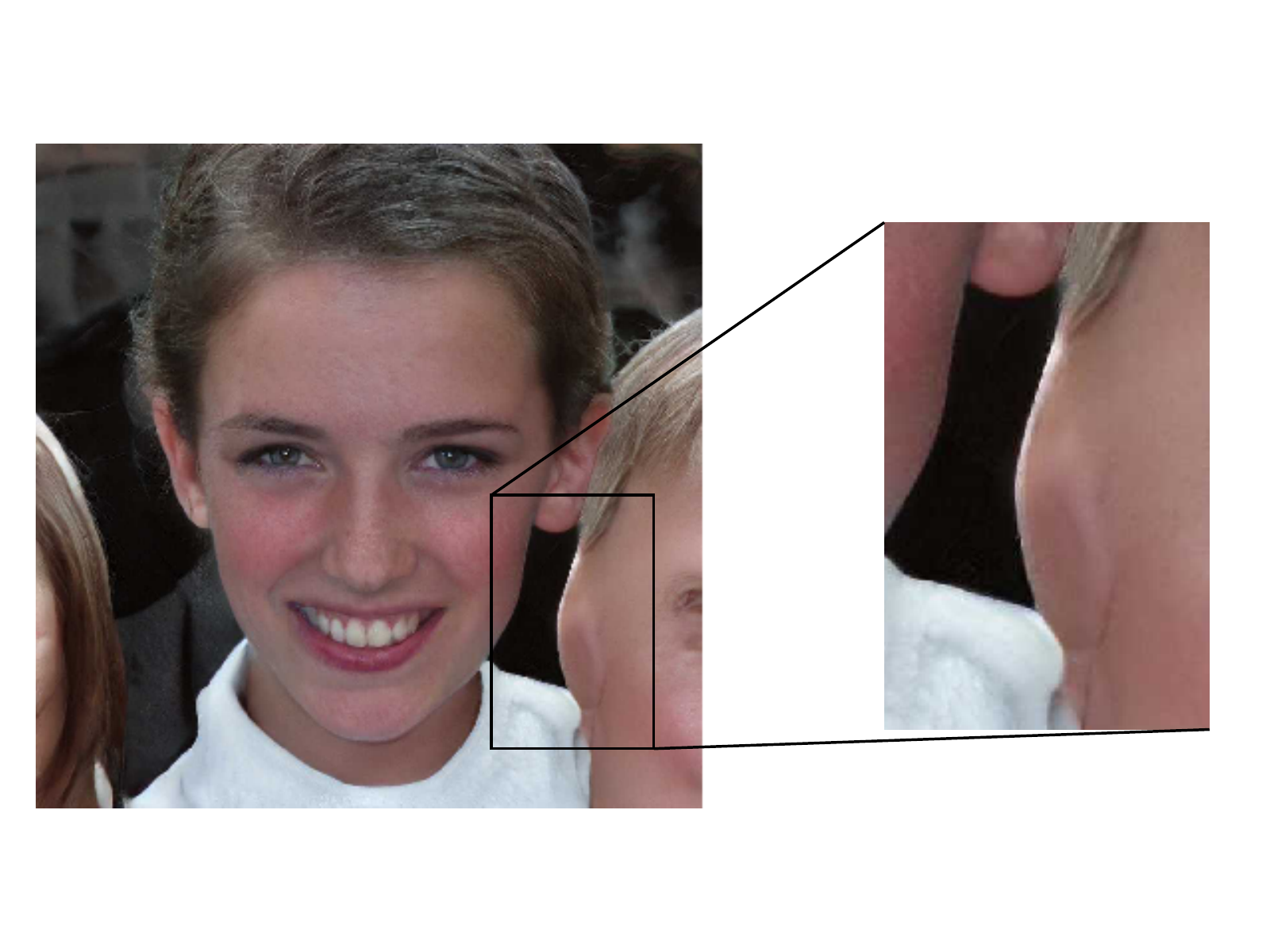}
  \caption{Synthetic face generated by StyleGAN2.}
  \label{fig:bias_example}
\end{figure}

In turn, \textit{attribution} comes down to attributing a depiction of a bias, \ie generated contents, back to the generator's bias. The bias can come in multiple forms. For example, \Cref{fig:bias_example} shows a StyleGAN2-generated face. The generator learned the characteristics of a human face and generates a highly-realistic synthetic face. However, we observe in the magnified insert that the generator produced an individual to the right of the artificial face and populated it with human-like features, yet failed to reconstruct the ear which should be in this location of the image. Similarly, it failed to reconstruct the neck of the individual to the left. In another instance, StyleGAN can often be recognised by droplet-like artifacts in the generated images~\cite{karras2019analyzing}. Bias can also take less recognisable format and come in the form of a frequency distribution~\cite{frank2020leveraging}. These are examples of \textit{biases} that \textit{attribution} attempts to capture in order to differentiate across different generators.

Capturing and attributing bias is something that humans inherently do---we are capable of attributing content in the domains of speech, music, art and even use of language. For example, a human art amateur is able to roughly determine the era in which a painting was created by observing the style of the drawing; with proper training, human experts can also attribute an art piece to its creator by authenticating signatures (such as strokes) that belong to individual artists. In a similar vein, here we aim to develop techniques to automatically capture biases of media generated by different GANs.

\section{Deepfake attribution}
\label{sec:relaxed_integrity}


\subsection{Threat Model}
\label{sec:threat_model}

We consider a \textit{white-box adversary} with knowledge of our approach. This means the adversary knows the details of the attribution mechanism introduced in \Cref{sec:reconstruction} and also the set $G$ of GANs we will consider as candidates for attribution. 
As far as adversarial capabilities are concerned, the adversary cannot control the sources of randomness (\ie the random initializations we bootstrap seed reconstruction with, or the style parameters we provide to generators). This is a standard assumption, similar to the ones  made in cryptography research: sources of randomness in GANs share some similarity to the concept of keys in cryptography.  

Adversaries can only control and manipulate the synthetic image which they intend to use as a deepfake for nefarious purposes.
We consider a series of adversaries with increasing ability to perturb deepfakes generated by a GAN to evade attribution. Choosing the perturbations we want our attribution to be robust to is not trivial because it involves the broader context our attribution mechanism will be deployed in. One could argue that only the original synthetic images should be correctly attributed to the model since any modification added to the image means that the image is no longer generated by a GAN exclusively. In this sense, the problem of robustness analysis should not exist as long as the attribution methods work well on original images. A similar problem arises in the copyright area where the call was left to the judges to determine whether the modification of the original work should be treated as fair use or copy right infringement \eg~\cite{wrightvwarnerbooks1991,cumminsvvella2002}. Nevertheless, we take a worst-case perspective and evaluate the robustness of our approach to unintended and adversarial manipulations of the deepfakes in addition to confirming that reconstruction works on original deepfakes. This is to demonstrate that the method can be useful in the real world to assist forensics experts investigating deepfakes created by adversaries attempting to defeat attribution, and despite the various transformations an image may go through online (\eg it may be compressed when uploaded to a social network platform). 
Specifically, we consider: 
\begin{itemize}
    \item Non-adversarial manipulations of the image applied as it is transmitted between its generation and attribution. This includes compression, cropping, and rotation.
    \item Adversarial modifications of the image (\ie adversarial examples) where adversaries actively seek to fool attribution. We use the Fast Gradient Sign Method~\cite{goodfellow2014explaining} and adapt the Carlini\&Wagner~\cite{carliniwagner} attack to our problem to produce deepfakes that increase the error of our seed reconstruction approach.
\end{itemize}




\subsection{Seed reconstruction}

\newcommand\mycommfont[1]{\footnotesize\ttfamily\textcolor{blue}{#1}}
\SetCommentSty{mycommfont}

\SetKwInput{KwInput}{Input}                
\SetKwInput{KwOutput}{Output}              
\begin{algorithm}[t]
\DontPrintSemicolon
  \KwInput{Synthetic image $\bf{x}$, set of generators $G = \{g_{0}, g_{1}, ... g_{n}\}$ and distance function $d$}
    
    \For{\bf{all} generators $g \in G$}
    {
        \For{$i \in \{1, \cdots, m\}$}
        {
            \tcc{Initial random seed selection}
            $\bf{s}$ = rand()\;
            \tcc{Randomness needed for generators}
            $R$ = rand()\;
            \tcc{Minimising distance of a reconstruction to a target image}
            $
            \begin{aligned}
            {\bf{s}'} = \operatorname*{argmin}_{{\bf{s}} \in \mathbb{R}^d} d(g({\bf{s}}, R), \bf{x})\\
            {{\bf{d}}[g][i]} = d(g({\bf{s}'}, R), \bf{x})\\
            \end{aligned}
            $\;
        }
        \tcc{Find the minimum distance among all seeds}
        $
        \begin{aligned}
        {{\bf{d}}_{g}} = \operatorname*{min}({{\bf{d}}[g]})\\
        \end{aligned}
        $\;
    }
    \tcc{Find the generator that corresponds to the minimum distance of all generators}
    $
    \begin{aligned}
    g' = \operatorname*{argmin}_{g \in G}{{\bf{d}}_{g}}\\
    \end{aligned}
    $\;
    \Return{$g'$}
\caption{Deepfake \textit{attribution} algorithm}
\label{alg:attribution}
\end{algorithm}

We assume that a given image can only be generated by a single GAN. Though this assumption may not be provable in theory, as we show in \S~\ref{sec:problem_def:limits} and Equation~\ref{eq:stylegan}, we find that this assumption works well in practice (see \S~\ref{sec:benign}) because the image is more likely to be regenerated by the original generator. Ideally, if the generator function were to be invertible, we could determine the seed and therefore claim attribution. 
However, this is not possible because deep neural networks are non-invertible. This is due to some of their architectural components: \eg activations such as the Rectified Linear Unit (ReLU) lose information (when their inputs are negative) and cannot be inverted. 
Researchers began investigating the design of neural architectures\textit{ for classification}, which are invertible by design~\cite{DBLP:journals/corr/abs-1811-00995}, but this work will not be applicable to generative models. 
Indeed, recall our discussion of generative modeling in \S~\ref{sec:pb-def}, a generative model maps a low-dimensional input (the seed) to a high-dimensional output (the synthetic image). This is unlike classifiers, which generally map a high-dimensional input (the image) to a low-dimensional output (the label). This suggests that the invertibility of generator functions faces additional challenges compared with classification. Thus, we focus on seed reconstruction without invertibility, as we discuss in the next section.


\subsection{Seed reconstruction algorithm}
Given a synthesized image $\bf{x}$ and a set of generators $G$, our algorithm reconstructs a seed for each generator that would lead it to synthesize $\bf{x}$, \textit{\ie} $\forall g \in G$ we search for a different $\bf{s}$ such that $g(\bf{s}) = \bf{x}$. 
Reconstruction here refers to a directed search through the latent space of a generator. The search is initialized with a random seed. An optimizer then uses the similarity between the $g(\bf{s})$, the synthesized image of the current seed, and $\bf{x}$, the target synthetic image, as a loss to direct the search over the space of possible seeds. We then compare the similarity of each synthesized image (from each seed) with the original image $\bf{x}$. Attribution is determined by the most similar synthesized image to the original image $\bf{x}$, where we use a distance function to determine similarity.


\subsubsection{Reconstruction function}\label{sec:reconstruction}
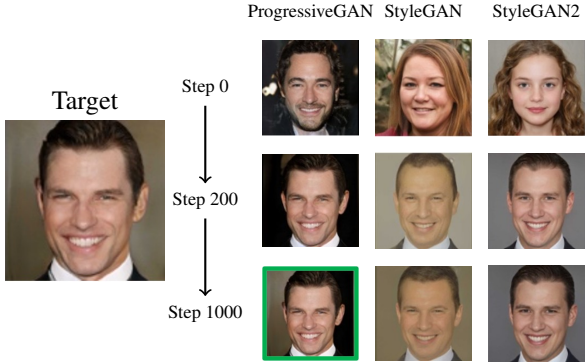
\begin{figure}[t]
\begin{tikzpicture}
\node[] at (0,3.3) {Target};
\node (A) at (1.6,3.5) {\scriptsize Step 0};
\node (B) at (1.6,2) {\scriptsize Step 200};
\node (C) at (1.6,0.5) {\scriptsize Step 1000};
\draw[->, thick]
  (A) edge (B) (B) edge (C);
\node[inner sep=0pt] (picture1) at (0,2)
    {\includegraphics[width=.25\linewidth]{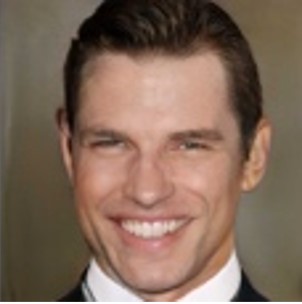}};
\node[] at (3,4.5) {\scriptsize ProgressiveGAN};
\node[inner sep=0pt] (picture2) at (3,3.5)
    {\includegraphics[width=.15\linewidth]{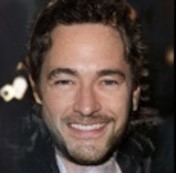}};
\node[] at (4.5, 4.5) {\scriptsize StyleGAN};
\node[inner sep=0pt] (picture3) at (4.5,3.5)
    {\includegraphics[width=.15\linewidth]{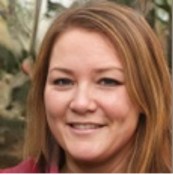}};
\node[] at (6, 4.5) {\scriptsize StyleGAN2};
\node[inner sep=0pt] (picture4) at (6,3.5)
    {\includegraphics[width=.15\linewidth]{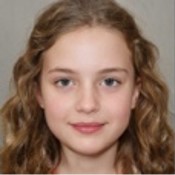}};
\node[inner sep=0pt] (picture5) at (3,2)
    {\includegraphics[width=.15\linewidth]{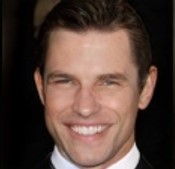}};
\node[inner sep=0pt] (picture6) at (4.5,2)
    {\includegraphics[width=.15\linewidth]{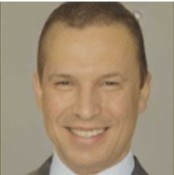}};
\node[inner sep=0pt] (picture7) at (6,2)
    {\includegraphics[width=.15\linewidth]{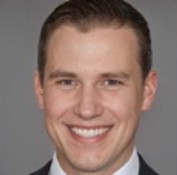}};
\node[inner sep=0pt] (picture8) at (3,0.5)
    {\includegraphics[width=.15\linewidth]{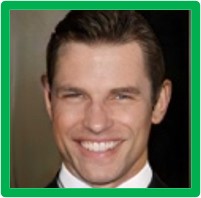}};
\node[inner sep=0pt] (picture9) at (4.5,0.5)
    {\includegraphics[width=.15\linewidth]{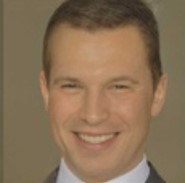}};
\node[inner sep=0pt] (picture10) at (6,0.5)
    {\includegraphics[width=.15\linewidth]{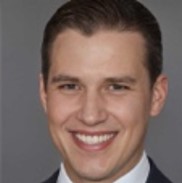}};
\end{tikzpicture}
\caption{Relaxed attribution approach reconstructing a deepfake.}
\label{fig:figure_0}
\end{figure}

The goal is to discover a seed that a generator could have used to generate the target synthetic image $\bf{x}$. Reconstruction can be formalized as finding a member of the preimage of $\bf{x}$ for $g$, \ie a seed $\bf{s}$ such that $g(\bf{s}) = \bf{x}$. 
We reformulate this into an optimization problem as follows:
\begin{equation}
\label{eq:reconstruction}
   {\bf{s}}_{g} = \operatorname*{argmin}_{\bf{s}} d({g}({\bf{s}}), {\bf{x}} ) 
\end{equation}
where $g(\bf{s})$ is the synthetic image recovered by generator $g$ and $d$ is a distance function to compare the recovered synthetic image to the target synthetic image $\bf{x}$. We explain how to choose a distance 
in \S~\ref{sec:distance_function}. The optimization problem in \Cref{eq:reconstruction} can be solved with gradient descent with respect to $\bf{s}$ because $g$ is a differentiable neural network so the overall distance computation is also differentiable. In practice, we use a variant of stochastic gradient descent commonly employed to optimize over neural networks: Adam~\cite{kingma2014adam}. We randomly initialize $\bf{s}$ and configure the optimizer to perform a fixed number of gradient descent steps (see illustration in \Cref{fig:figure_0}).  We make two observations: (i) we are not trying to reconstruct the original seed $\bf{s}$ that generated the synthetic example $\bf{x}$, which is shown later to be difficult (refer\S~\ref{sec:benign}), and (ii) we may obtain multiple reconstructed seeds if we run our optimization algorithm multiple times with a different random initialization. In the following, we refer to each of these runs as a \textit{reconstruction attempt}.
We record the final reconstruction distance $d({g}({\bf{s}}), {\bf{x}} )$, which is used later to infer attribution.


\subsubsection{Distance Function}\label{sec:distance_function} The distance function $d$ in \Cref{eq:reconstruction} measures the difference between the recovered synthetic image and the target synthetic image. The choice of  distance function $d:\mathbf{X} \times \mathbf{X} \longrightarrow \mathbb{R}$ is two fold: (i) the function should be differentiable and favor numerically stable symbolic gradient computations so we can optimize over a smooth loss surface, and (ii) the function should capture semantics of the images and reflect human perception as much as possible.  


We introduce an effective distance function as \textit{$\ell_2$-feature}. The function is defined as $||f({\bf{x}}) - f(g(\bf{s}))||_2$ where $f$ is a feature extraction function: we extract a feature map from both the reconstructed and the target image, before computing an $\ell_2$ distance over these features. We utilize one part of a pretrained Inception-V3~\cite{szegedy2015rethinking} network to extract feature maps. Starting from 299$\times$299 inputs, we extract the Inception-V3 network from the first inception block (referred to as \textit{Mixed\_5d} layer~\cite{tensorflow_slim}), which outputs a 35$\times$35$\times$288 feature map for each input. This layer balances feature abstraction and granularity. When necessary, we bi-linearly resize images produced by the generator to match the feature extractor's input size.

\subsubsection{Attribution} Once we obtain a set of reconstructed seeds for each candidate model $g$ in the set of generators $G$, we need to compare these seeds to attribute the  synthetic image $\bf{x}$ to one of the generators. We need to ensure that the comparison is done over quantities that have comparable range and dimensionality. This ensures that we do not favor any of the generators due to their architecture. Hence, we use the final reconstruction distance achieved by each generator, $d(g({\bf{s}}_g), {\bf{x}})$, where ${\bf{s}}_g$ is the seed returned by the reconstruction process as defined in \Cref{eq:reconstruction}. When we run multiple reconstruction attempts for the same model, we only consider the attempt that led to the seed ${\bf{s}}_g$ with the smallest reconstruction distance for the generator. Taking the minimum rather than averaging over all reconstruction attempts ensures that we are more robust to outliers caused by optimization imperfections resulting from the non-convex nature of the problem. Thus, we attribute ${\bf{x}}$ to generator $g_{\bf{x}}$ according to:
\begin{equation}
    \label{eq:attribution}
    g_{\bf{x}} = \operatorname*{argmin}_{g \in G} d(g({\bf{s}}_g), \bf{x}) 
\end{equation}
We also considered recording ${\bf{s}}, {\bf{x}}, {d}({g}({\bf{s}))}$ for each intermediate step of every reconstruction attempt and using this information to further enrich the attribution decision. 
In practice, we however find that selecting the model with the smallest reconstruction distance at the end of the reconstruction as the attributed model for a target image is simpler and sufficient.


 
\subsubsection{Evaluating Attribution} The performance of our relaxed attribution algorithm can be evaluated by generating a dataset of synthetic images. In \S~\ref{sec:implementation} and~\ref{sec:eval}, we design a dataset where we synthesize images using three GANs for face generation. Because we generate the images ourselves, we label them with the ground truth attribution. We then  evaluate the performance of our relaxed attribution algorithm by calculating the accuracy between this ground truth label and the output of our algorithm.  We also report confusion matrices to visualize the relative performance of attribution on each GAN. 

\section{Dataset collection and experimental setup}
\label{sec:implementation}

Before we evaluate our approach, we created a dataset from modern GAN architectures capable of generating high-quality and high-resolution images of human faces. We release our experimental code here~\footnote{\url{https://anonymous.4open.science/r/e86c8def-bef4-4af9-8938-f6dc9774cf2f/}} to facilitate reproducibility of our results, and comparison with follow-up work on attribution. 

\subsection{Dataset collection} To obtain the synthetic images we need for our experiments, we utilize three state-of-the-art generative models (GANs):
\begin{itemize}
    \item \textit{ProgressiveGAN}~\cite{karras2017progressive} trained on CelebA-HQ data~\cite{karras2017progressive} produces synthetic human faces with progressively growing resolution up to 1024$\times$1024 pixels. 
    \item \textit{StyleGAN}~\cite{karras2019style} and \textit{StyleGAN2}~\cite{karras2019analyzing} compose networks that map random seeds into latent vectors before feeding them as \textit{styles} to subsequent layers to exert specific control over attributes of the synthesized outputs. Both GANs are trained with the FFHQ dataset~\cite{karras2019style}.
\end{itemize}
Our implementation of these models is based on the official implementation provided by the authors of each paper in a public repository. We also use the pre-trained weights provided with the code to ensure that each GAN architecture was trained exactly as the authors intended. We note that  the seed vector's dimensionality is 512 for all models, where each vector component is a real number. 


Using each of the three models described above, we generate 2000 synthetic faces for a total of 6000 faces. Each image has 1024$\times$1024 RGB pixels with a human face in the center as illustrated in \Cref{fig:faces_11}. For StyleGAN and StyleGAN2, we set the truncation rate to 0.7. The truncation rate balances the diversity and quality of generated images by rescaling the deviation of feature values from its center; a value closer to 1 will result in more diverse, but lower quality images. 




\subsection{Experimental setup} 

For each of the 6,000 images in our dataset, we attempt reconstruction (ergo attribution) with all 3 generative models. For each of the corresponding 18,000 seeds to be reconstructed, and we attempt this reconstruction with three different random initializations for each of the 18,000 seeds to be reconstructed. This results in 54,000 reconstruction attempts. 

We use the attribution approach described in \S~\ref{sec:reconstruction}. In our experiments, we used the Adam optimizer to minimize the objective in Equation~\ref{eq:reconstruction} and set its learning rate to 0.1, with no explicit learning rate decay or schedule.  For each target image and its 9 reconstructions (three models and three random seeds per model), we only retain the reconstructed seed which achieved the smallest reconstruction distance to determine attribution as described in \S~\ref{sec:reconstruction}.

In line with our threat model from \S~\ref{sec:threat_model}, we evaluate the attribution process against adversaries applying perturbations that make it increasingly more difficult to attribute the resulting modified deepfake:
\begin{enumerate}
    \item Benign setting: here, we attribute images directly synthesized by a generator and left unperturbed. 
    \item Non-adversarial modifications: we consider compression, cropping, and rotations. These transformations are commonly encountered by media in transmission across different platforms. Some of the modifications are unbounded in the sense that they apply large changes to the image (as measured by an $\ell_p$ norm)
    \item Norm-bounded adversarial examples: we  use the Fast Gradient Sign Method~\cite{goodfellow2014explaining} and adapt the  Carlini\&Wagner~\cite{carliniwagner} attack, to adversarially modify the deepfake within an $\ell_p$ norm bound. We modify the attack objective to maximize the reconstruction distance rather than the cross-entropy as done in the attacks originally proposed by the original authors. Here, we study worst possible scenario for attribution -- attribution is considered in presence of the strongest adaptive White-box attacker. 
\end{enumerate}

Our implementation is written in TensorFlow with Python3.7. Due to the large computational cost, we distribute experiments on a cluster with hundreds of GPUs: each reconstruction attempt runs independently in a job on one NVIDIA T4 GPU with three CPU cores and 10GB of RAM. 
\section{Evaluation of seed reconstruction}
\label{sec:eval}



\subsection{Summary}
The objective of our evaluation is to understand the efficacy of the relaxed integrity approach. From our analysis, we draw the following insights on the performance of the relaxed attribution method: 

\begin{enumerate}
\itemsep0em
\item We observe that in a fully benign setting, the relaxed attribution method can successfully attribute 97.62\% of 6000 images with \textit{as few as} three seed reconstruction attempts (\ie different random initializations) per image. Attribution accuracy increases with the number of reconstruction attempts (see \S~\ref{sec:benign}).
\item We observe that image manipulations that result in small structural changes (\eg JPEG compression) do not impact accuracy of our approach. However, manipulations that induce large changes (\eg mirroring an image) negatively impacts attribution accuracy unless they are considered during  attribution (see \S~\ref{sec:no_adv_manip}).
\item We observe that norm-bounded adversarial examples negatively impact attribution accuracy when the perturbation is significantly visible to human eyes (see \S~\ref{sec:adv_manip}).
\item We observe that our method can successfully attribute images generated from very similar generators. We find that fine-tuned models can be successfully attributed with 90.9\% accuracy, as opposed 97.62\% for unrelated models\footnote{Additionally, when the ground truth model is not present, relaxed attribution method can attribute images to another model with the same architecture with 98\% accuracy (see \Cref{sec:finetune_notrue}).}.
\item We conduct a user study and confirm that humans can attribute reconstructed images to the target deepfake with high accuracy (89.81\%). Our relaxed attribution algorithm agrees with human judgement 93.7\% of the time (see \S~\ref{sec:user_study}).
\end{enumerate}

\subsection{Benign Setting}
\label{sec:benign}

\begin{figure}[t]
    \centering
    \includegraphics[scale=0.35]{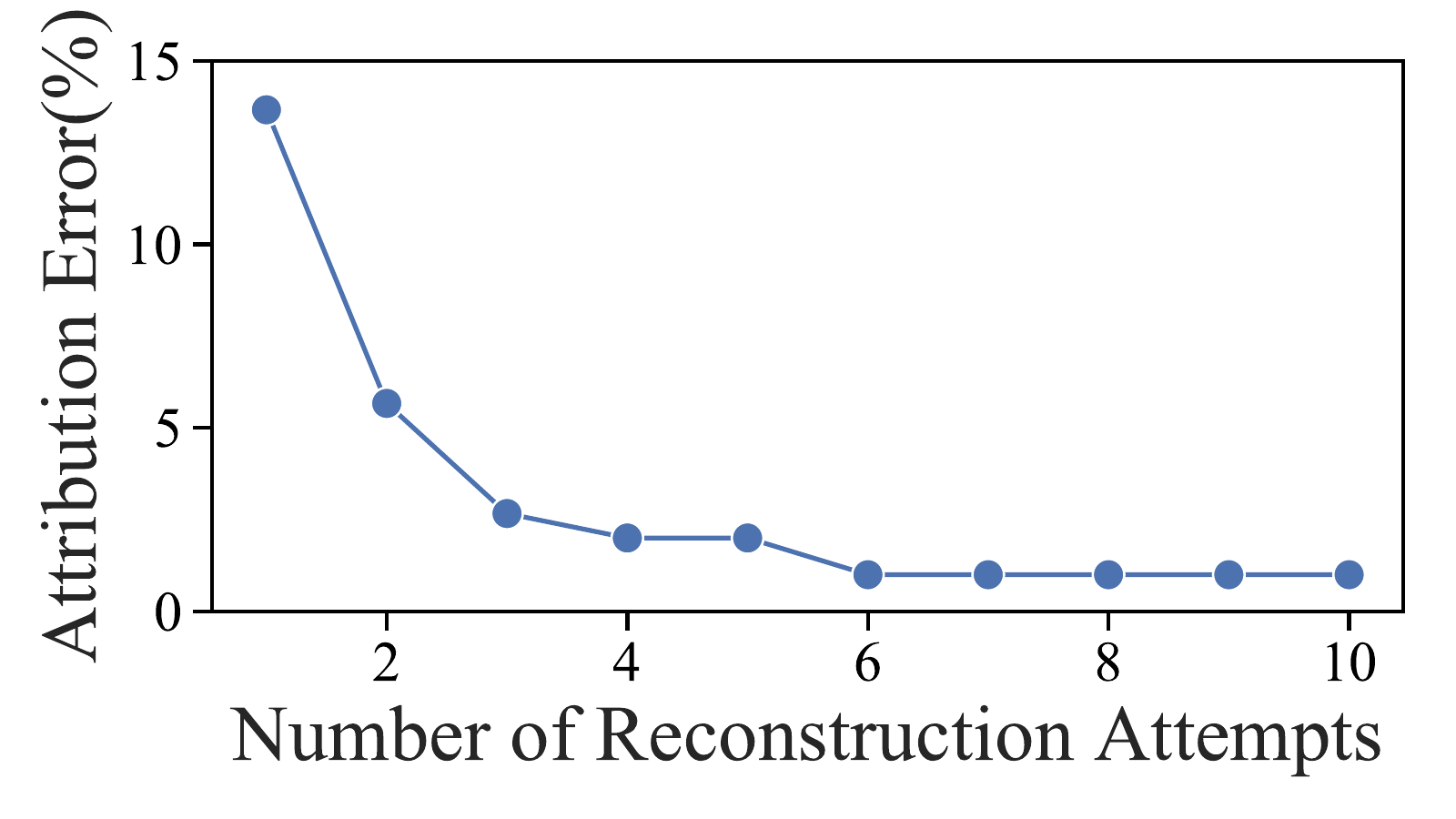}
    \caption{Increasing the number of reconstruction attempts decreases the attribution error, before it plateaus.}
    \label{fig:error_v_n_repeats}
\end{figure}

Recall that for each synthetic image, we verify if the image is generated by one of the 3 candidate GANs. Confusion matrices are shown in Figure~\ref{fig:confusion_matrices}. Out of 6000 images, 143 are attributed to the wrong GAN, which represents an error rate of 2.38\%. Out of the 143 failure cases, 120 of them are StyleGAN synthesized images, 22 StyleGAN2 generated, and 1 ProgressiveGAN generated: attribution is more difficult  on certain GANs. A visual inspection led us to find that most failure cases were due to reconstructions being stuck at a local minima on the correct GAN for all reconstruction attempts.  An example of a diverged optimization is shown in Figure~\ref{fig:face_43}. With more reconstruction attempts (\ie with more random initializations for the seed reconstruction algorithms), this  can be mitigated, as shown in Figure~\ref{fig:error_v_n_repeats}. Interestingly, we find that there is no bijection between generator's synthesized outputs and the seeds used to generate them. We discuss it in more details in~\Cref{sec:seed_difference}.







\subsection{Non-Adversarial Manipulations}
\label{sec:no_adv_manip}
Manipulations discussed here are those that are not intentionally made to avoid detection/attribution. 

\noindent{\bf 1. JPEG Compression: } Images circulating online can often be converted into different formats for efficient distribution and storage. This process is lossy and often introduces perturbations. To evaluate our method's performance in such cases, we utilize JPEG compression. Specifically, we compress target images using OpenCV with varying compression ratios (50\%, 70\%, 90\%, 100\%), before using them as input to our attribution methodology. Our algorithm achieved an error rate of 4.3\%, 4.6\%, 4.6\%, 4\% respectively, which shows that JPEG compression has little impact on attribution accuracy. This could be explained by the down-scaling operation before feature extraction during the reconstruction procedure, which effectively reduces any artifacts introduced by  compression. 



\noindent{\bf 2. Image Augmentation:} Apart from compression, we apply several image augmentation strategies obtained from the open-source \texttt{albumentations} library~\cite{info11020125}. We list them all in Table~\ref{table:augmentation}. Note that these augmentations do not change the semantics of the image. We summarize some salient observations:


\begin{figure}[t]
    \centering
    \includegraphics[width=0.65\columnwidth]{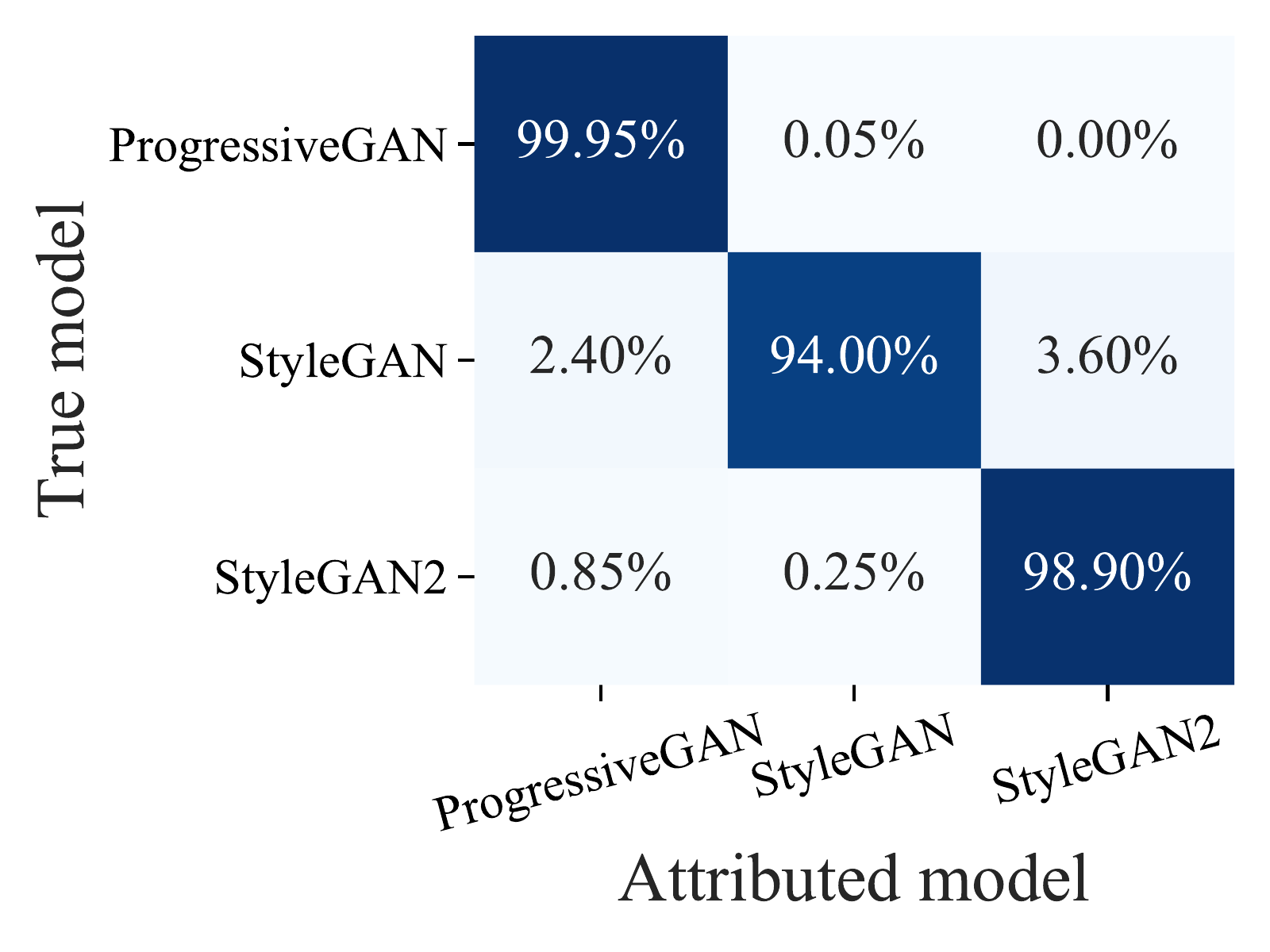}
    \caption{Normalized confusion matrix on 6,000 images, 2,000 from each GAN. The overall error rate is 2.38\%.}
    \label{fig:confusion_matrices}
\end{figure}

\begin{table}[t]
  \centering
  \resizebox{\columnwidth}{!}{
  \begin{tabular}{cccc}
    \toprule
    \cmidrule(r){1-2}
    {\bf Augmentation}     & \begin{tabular}[x]{@{}c@{}}{\bf Error Rate}\\(3 seeds)\end{tabular}   & \begin{tabular}[x]{@{}c@{}}{\bf Error Rate}\\(10 seeds)\end{tabular} & {\bf Description} \\
    \midrule
    \midrule
    Gaussian Blur &5\% &2.3\% & \begin{tabular}[x]{@{}c@{}}Gaussian blur with kernel size \\ between 3 - 7\end{tabular}\\
    Gaussian Noise & 12.3\% &9\% & Gaussian Noise $\mathcal{N}(0,\,0.01)$\\
    Mirror & 25.3\% &23\% & \begin{tabular}[x]{@{}c@{}}Flip each pixel's x axis long the \\ center line\end{tabular}\\
    Random Crop & 14.3\% &9.7\% & \begin{tabular}[x]{@{}c@{}}Crop image on both axes between \\ 100\% - 90\%\end{tabular} \\
    Random Rotate & 5.7\% &3\% & Rotate image between 0-5 degrees\\
    Zoom In    & 10.3\% &5.7\% & Crop the center 90\% on both axes  \\
    \bottomrule
  \end{tabular}
  }
\caption{Augmentations for non-adversarial manipulations}
  \label{table:augmentation}
  \end{table}

\begin{itemize}
\itemsep0em
\item Gaussian Noise and Random Rotate may leave noticeable artifacts indicating the image has been augmented. However, as stated in previous sections, repeating the attribution process with more random initializations to reconstruct the seed will increase success rate. 
\item Mirroring an image significantly decreases the attribution's accuracy (to less than 75\%): GANs are unable to reconstruct asymmetric faces. This effect is particularly pronounced on the StyleGAN model. 


\end{itemize}


\begin{figure*}[t]%
    \centering
    \subfloat[Adversarial perturbation (FGSM) in the seed space: $\varepsilon$ is $\ell_\infty$ bounded. \label{fig:error_v_eps_latent}]{\centering
    \includegraphics[width=0.6\columnwidth]{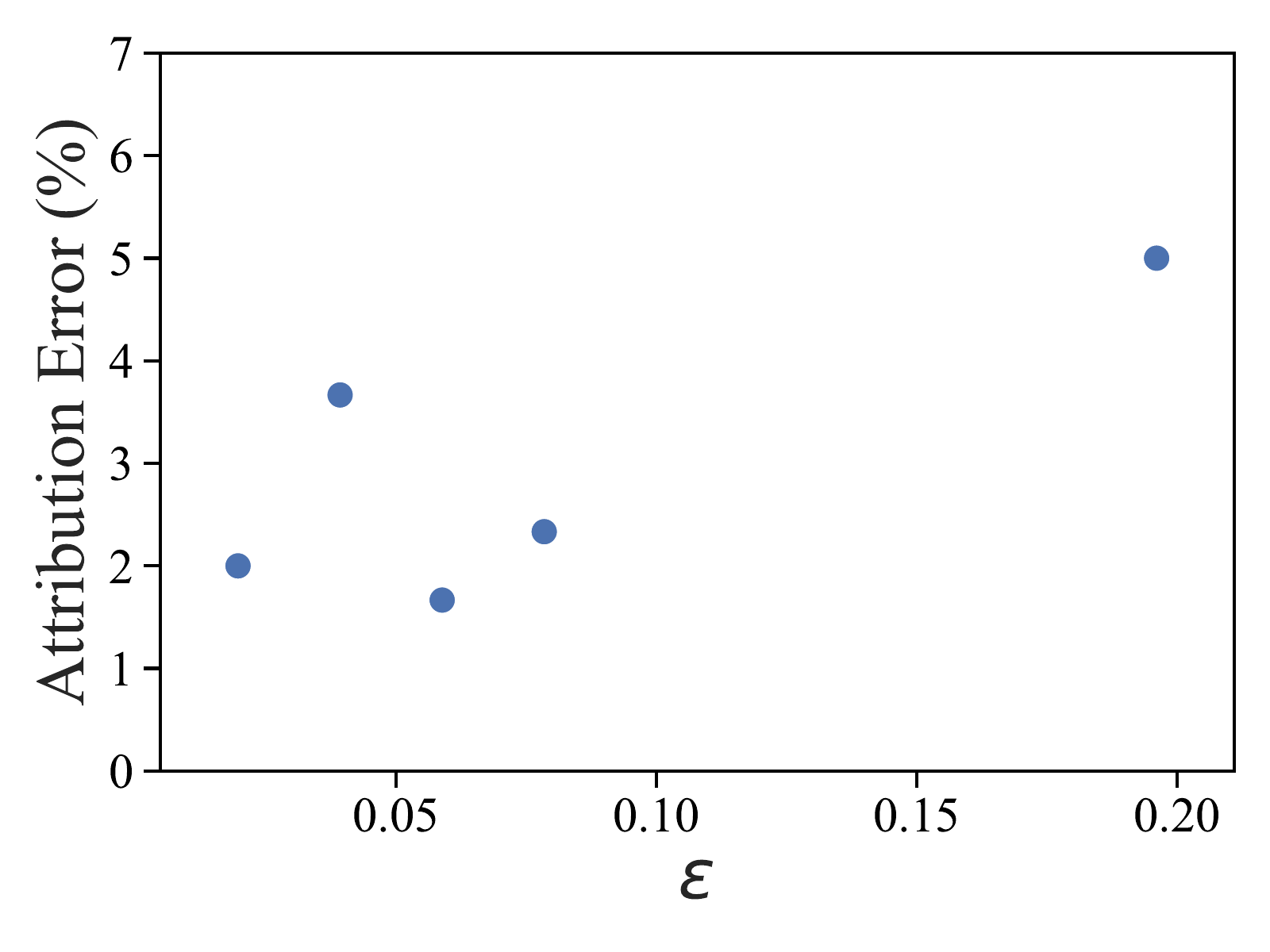}}\hfill
    \subfloat[
    Adversarial perturbation (FGSM) in the image space: $\varepsilon$ is $\ell_\infty$ or $\ell_2$ bounded.
    \label{fig:error_v_eps_image}] {\includegraphics[width=0.6\columnwidth]{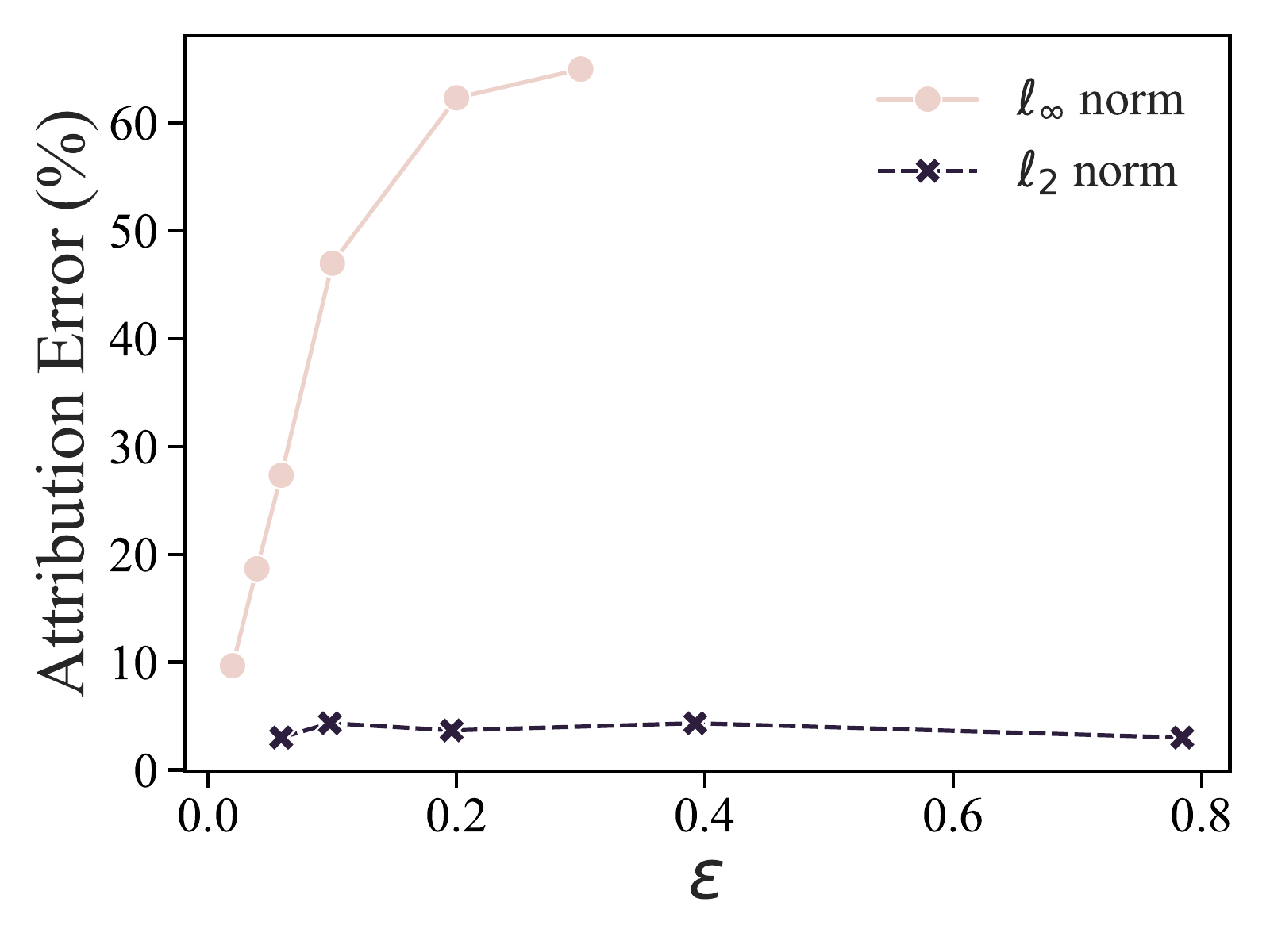}}\hfill
    \subfloat[
    Adversarial perturbation (CW) in the image space: $\delta$ is $\ell_2$ bounded. Numbers plotted in the figure represent values of $c$. 
    \label{fig:error_v_c_carlini}]{\includegraphics[width=0.6\columnwidth]{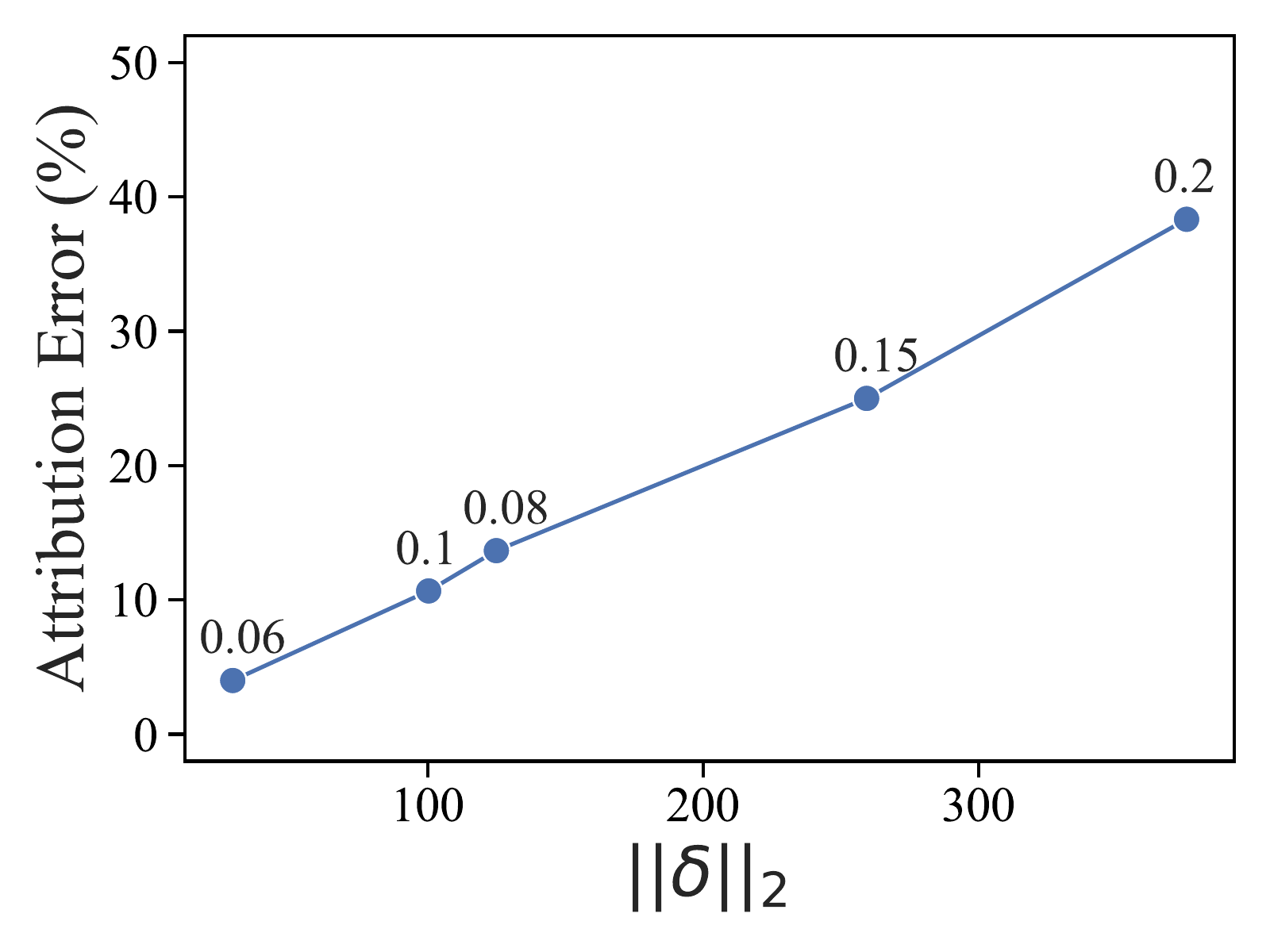}}

    \caption{Performance of relaxed attribution algorithm against three types of adversarial perturbations (denoted $\varepsilon$ or $||\delta||_2$).}
    \label{fig:adv_eval}
\end{figure*}

\subsection{Adversarial Manipulations}
\label{sec:adv_manip}

The previous section details manipulations made to deepfakes without the explicit intent of having the attribution process fail. In this section, we thus focus on adversarial perturbations. Recall that the attribution process has two steps. The first is to reconstruct the image given multiple seed initializations and generator architectures, and the second to choose the generator whose reconstructed image is closest to the synthetic image under consideration. We attack both steps in our evaluation: the first attack below (seed space) seeks to make it harder for the optimization process to converge, the second (image space) manipulates the distance metric used for attribution, whereas the third attack crafts images that are difficult to attribute by treating the overall attribution process as a black-box classifier. To be conservative, we analyse robustness in a white-box setting with full knowledge of both the generative models and the feature extractor used to measure the distance $d$ in \Cref{eq:reconstruction}.

\noindent{\bf 1. Adversarial Perturbations in the Seed Space:} Adversaries may try to find seeds that generate hard-to-reconstruct images. If successful, images produced by such seeds can not be effectively reconstructed on the original model despite the absence of any further manipulations. 
We use the FGSM method to generate adversarial latent seeds: the adversary considers the generator and composes it with the feature extractor, treating the two as a single model. The adversary can then perturb the seed (under an $\ell_\infty$ constraint) to form an adversarial seed which leads the generator to output a synthetic image that exhibits a feature representation far from the original deepfake.  

\begin{figure}[t]
\centering
\subfloat[Generated with FGSM on the seed space with $\varepsilon$ = 0.0169.]{\label{fig:adv_latent_05_target_0}{\includegraphics[width = 0.32\linewidth]{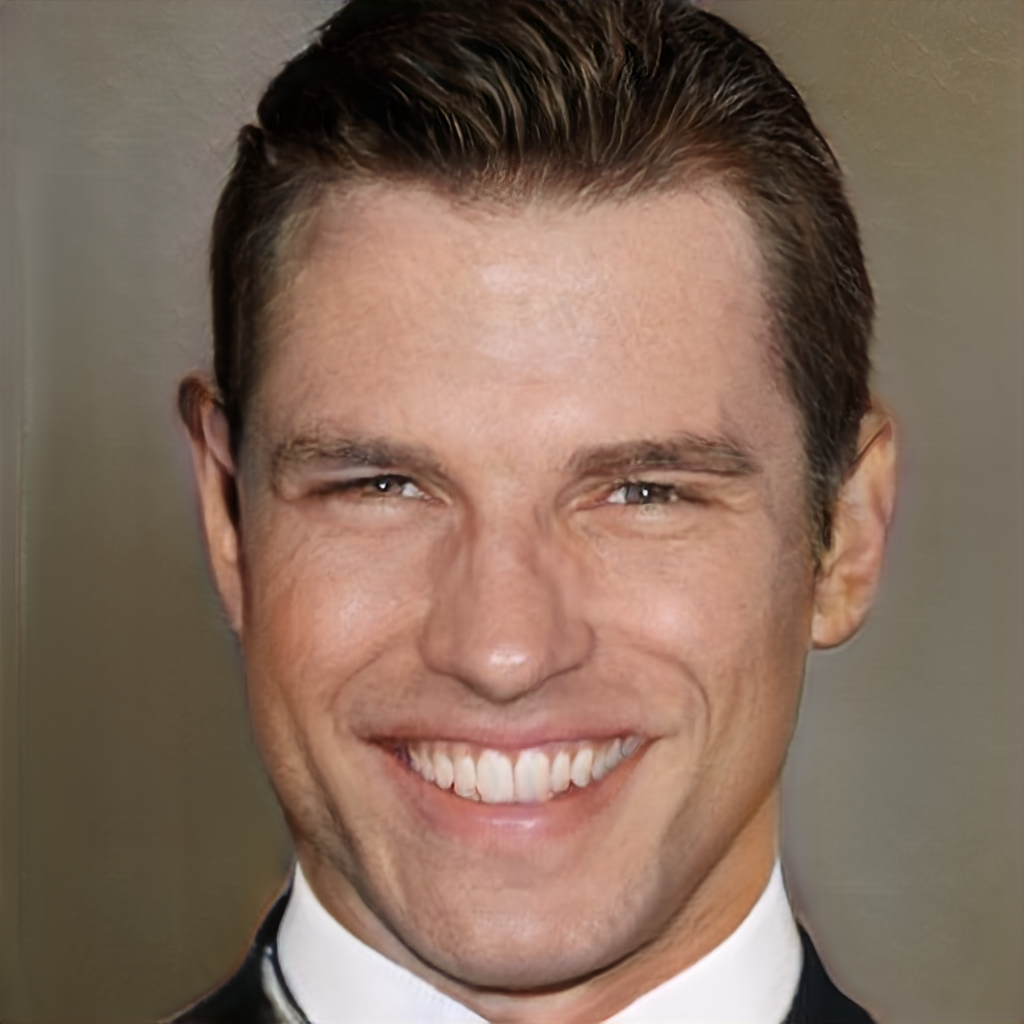}}  } \hfil
\subfloat[Generated with FGSM on the seed space with $\varepsilon$ = 0.039.]{\label{fig:adv_latent_010_target_0}{\includegraphics[width = 0.32\linewidth]{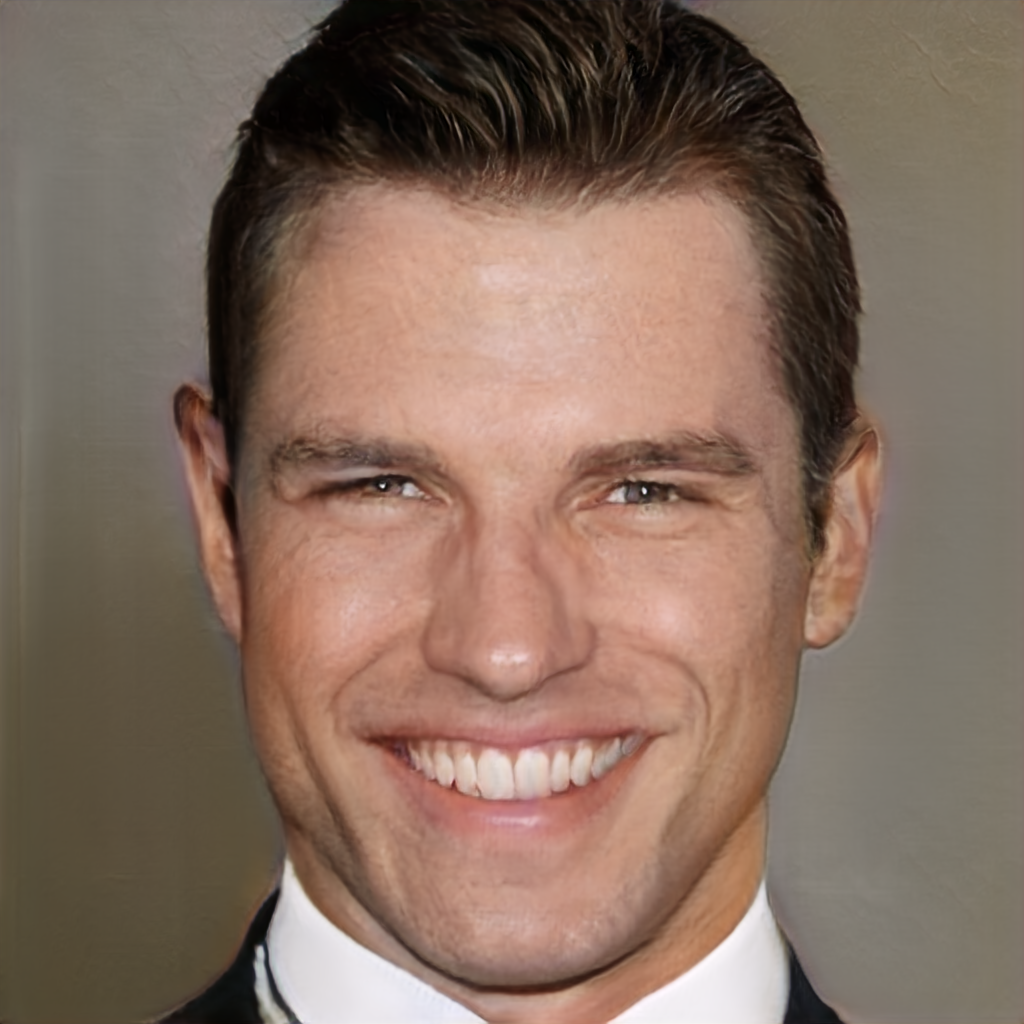}}  } 

\subfloat[Generated with FGSM on the seed space with $\varepsilon$ = 0.078.]{\label{fig:adv_latent_020_target_0}{\includegraphics[width=0.32\linewidth]{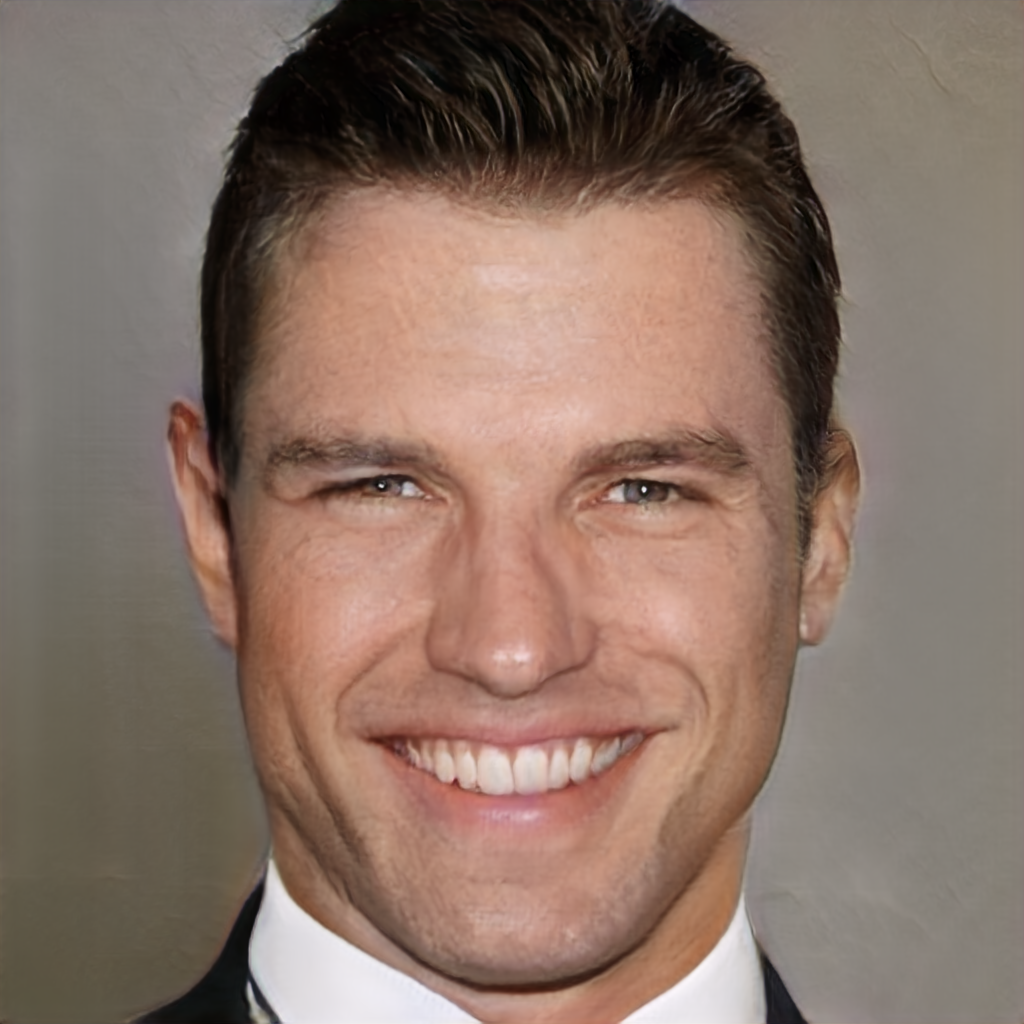}}  } \hfil
\subfloat[Generated with FGSM on the seed space with $\varepsilon$ = 0.196.]{\label{fig:adv_latent_050_target_0}{\includegraphics[width=0.32\linewidth]{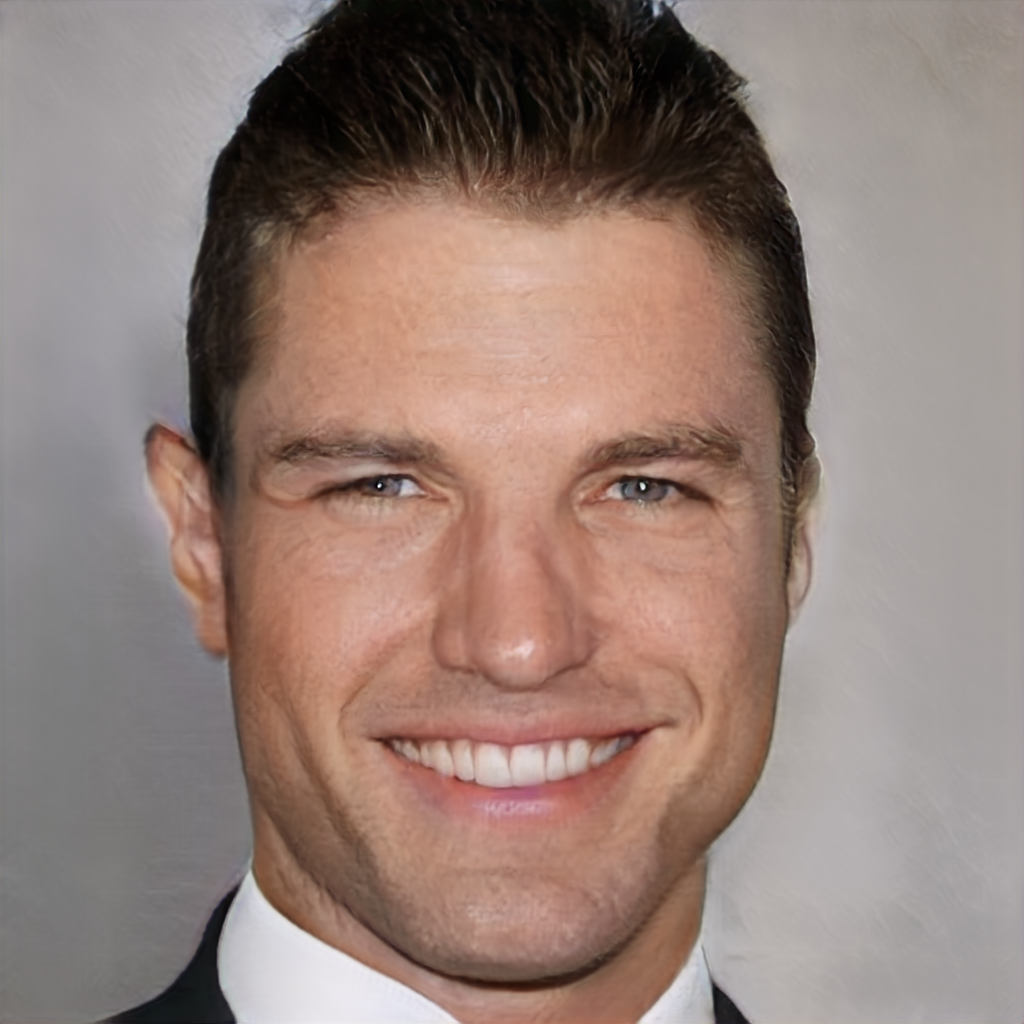}}  } 
\caption{Adversarial example generated by FGSM on seed space. The character in display has significantly changed with increasing $\varepsilon$. }
\label{fig:adv_seed_example}
\end{figure}

Results are shown in Figure~\ref{fig:error_v_eps_latent}. We find that this strategy is not very effective. Because the adversary is perturbing the seed and not the image, this makes it easier to find an adversarial seed which does not result in noticeable artifacts in synthetic images. This however comes at the expense of introducing artifacts that alter the high level characteristics of the image: Figure~\ref{fig:adv_seed_example} shows an example image that gradually changes its characteristics as the adversarial perturbation on the seed space increases in magnitude (as measured by the $\ell_\infty$ norm of $\varepsilon$). 
We did not find that switching to a more powerful optimization procedure such as the one used in the CW attack helps. 
For this reason, we consider next an attack in the image space. 




\begin{figure}[t]
    \centering
    \subfloat[Generated with FGSM ($\ell_\infty$) with $\varepsilon$ = 0.0588. Attribution error = 0.27. SSIM = 0.874.\label{fig:adv_015_target_0}]
    {\includegraphics[width=0.46\linewidth]{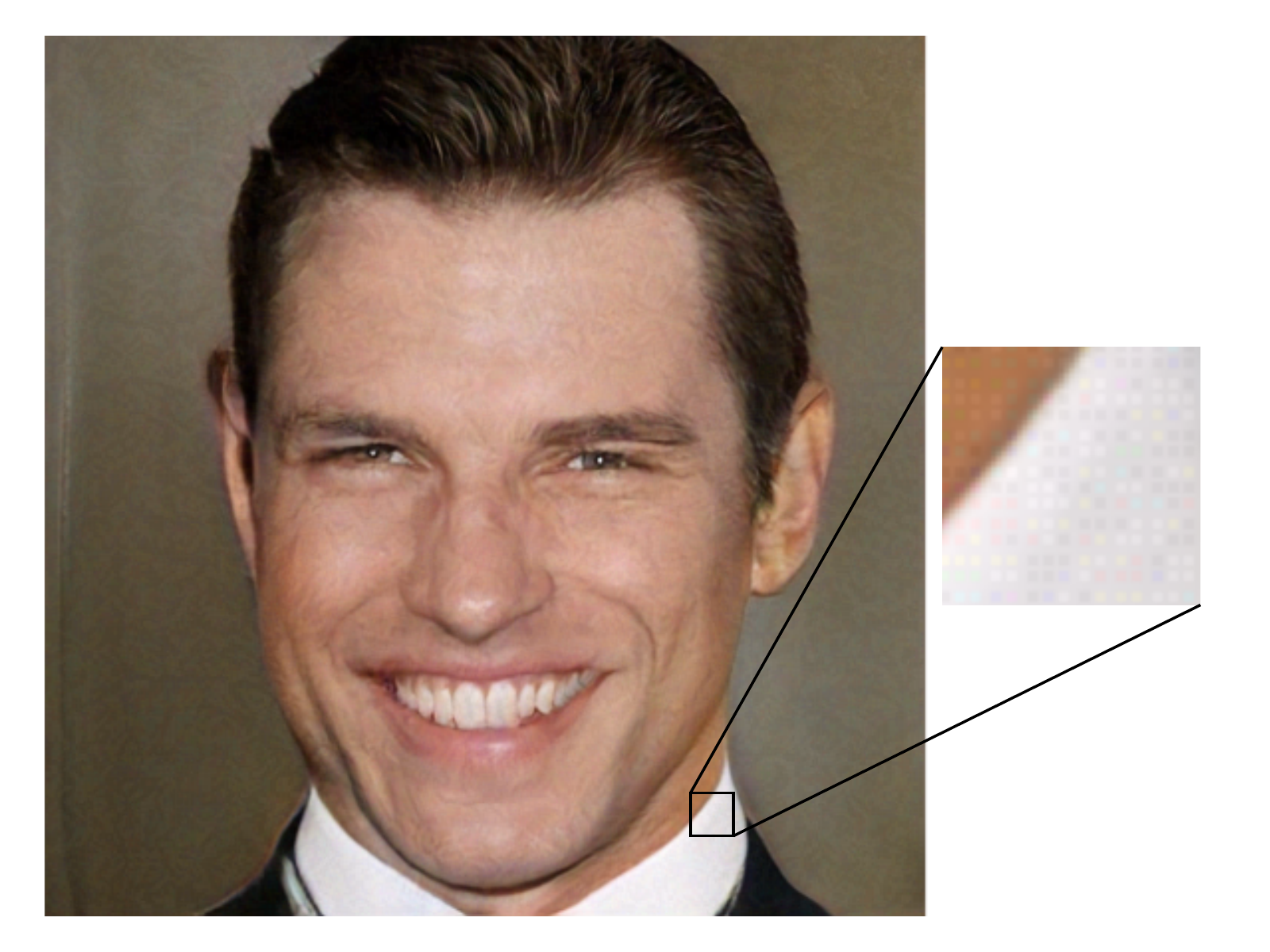}}\hfil
    \subfloat[Generated with FGSM ($\ell_\infty$) with $\varepsilon$ = 0.1. Attribution error = 0.47. SSIM = 0.719.\label{fig:adv_01_taraget_0}]
    {\includegraphics[width=0.46\linewidth]{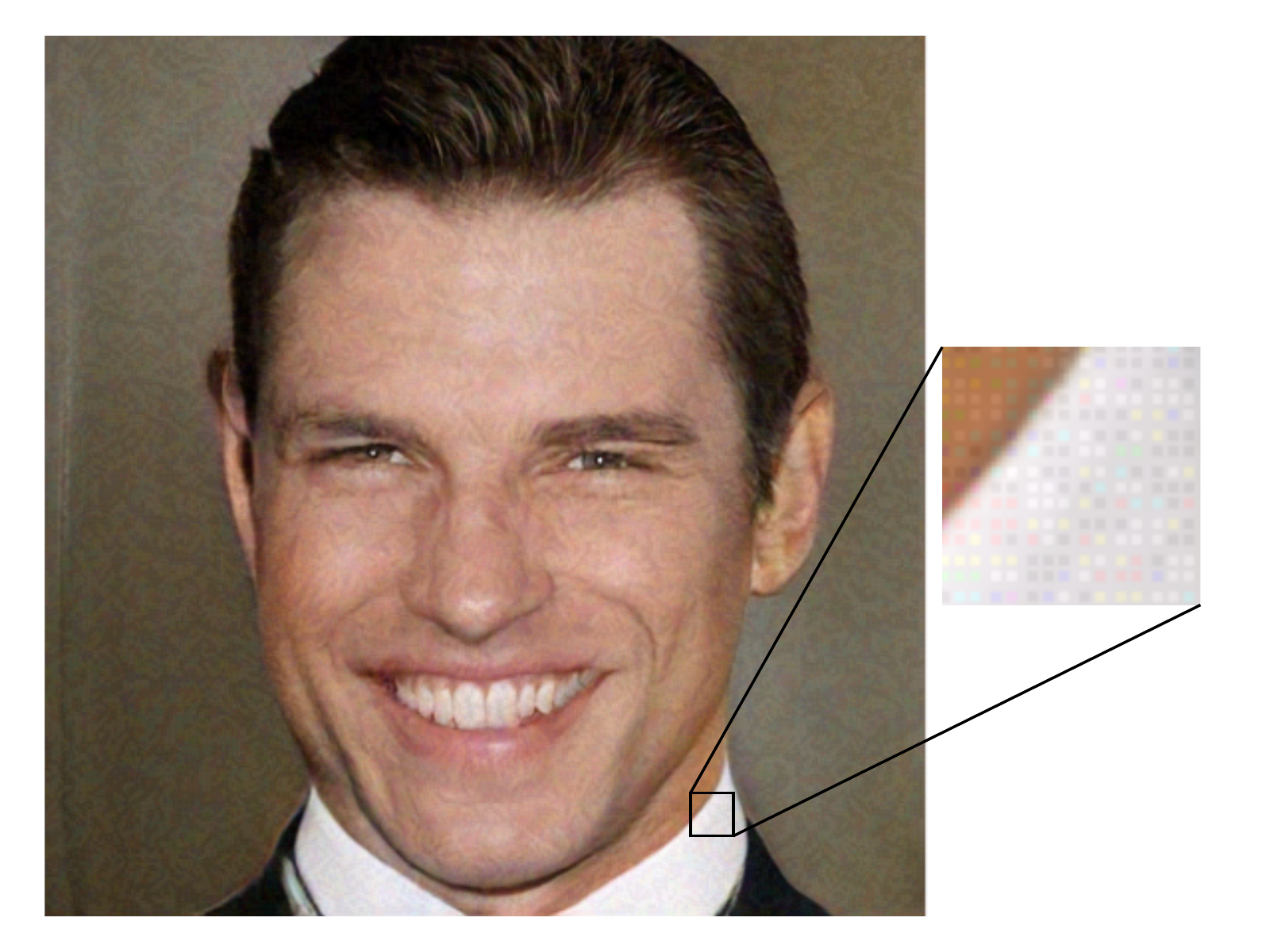}}
    
    \subfloat[Generated with adapted CW ($\ell_2$) with $||\delta||_2$ = 100. Attribution error = 0.11. SSIM = 0.883.\label{fig:adv_carlini_01_target_0}] {\includegraphics[width=0.46\linewidth]{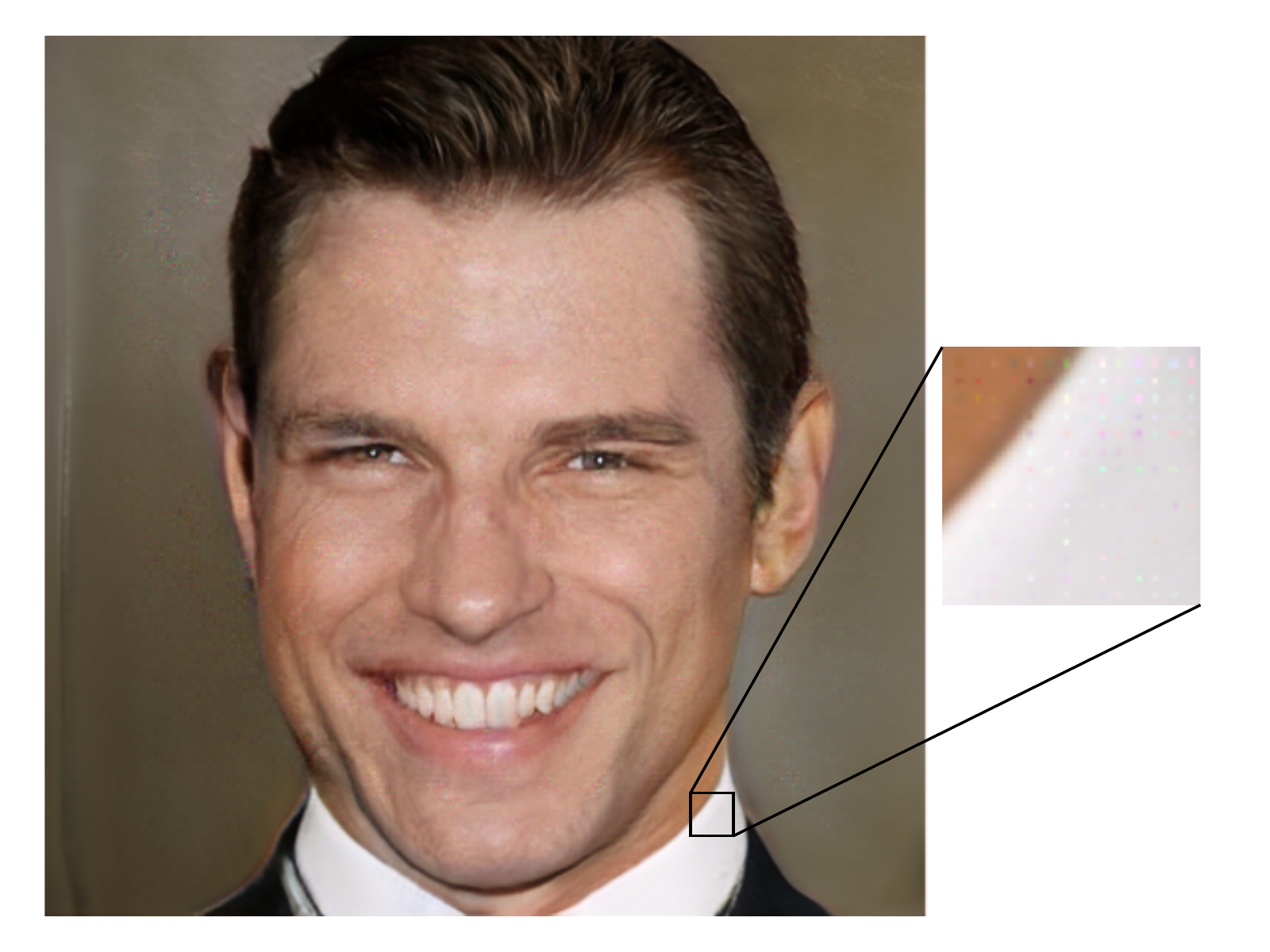}}\hfil
    \subfloat[Generated with adapted CW ($\ell_2$) with $||\delta||_2$ = 260. Attribution error = 0.25. SSIM =0.657.\label{fig:adv_carlini_015_target0}]{\includegraphics[width=0.46\linewidth]{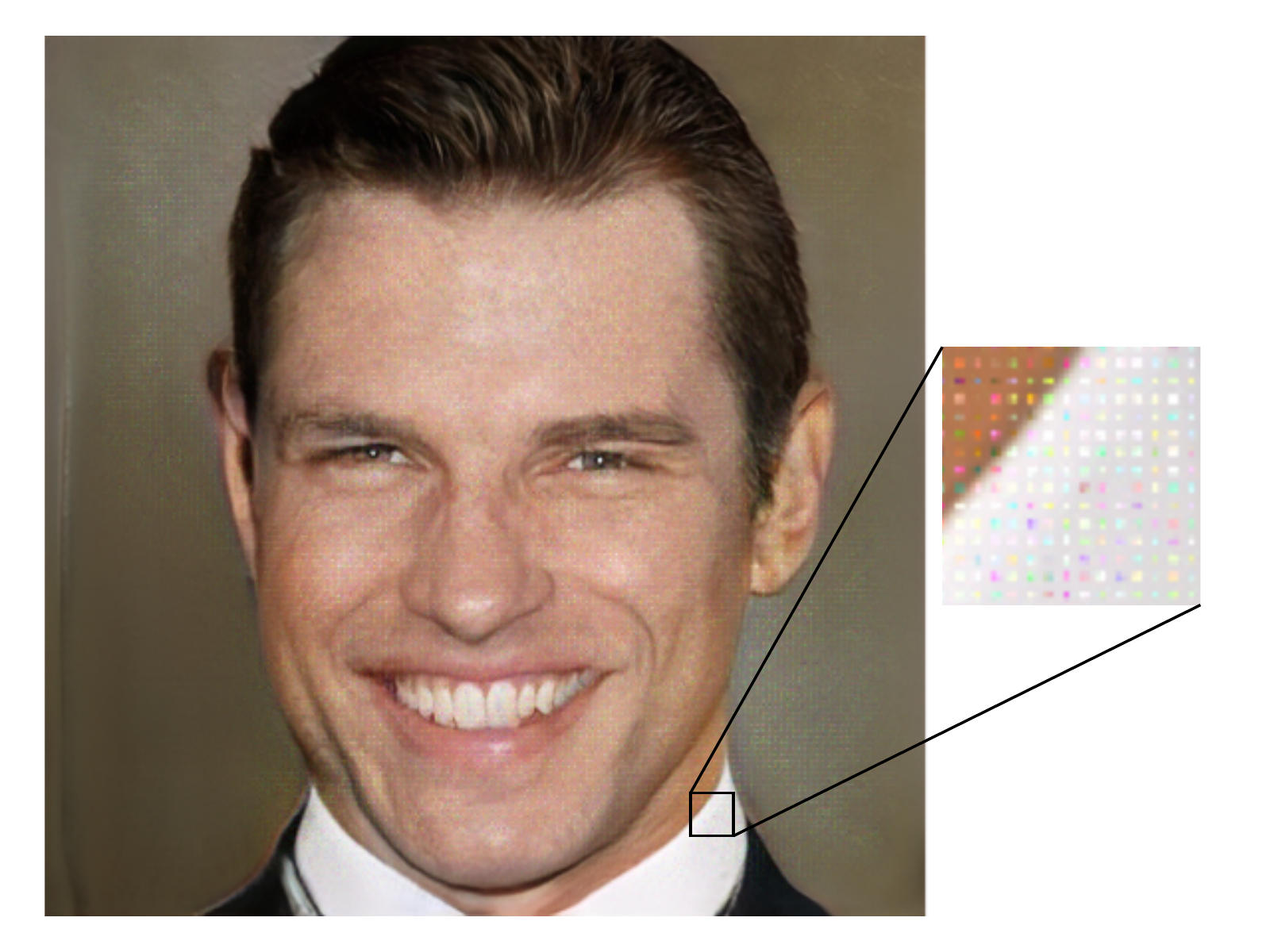}}\hfil
    \caption{Adversarial examples generated by FGSM and adapted $\text{CW}$ perturbation, both in the image space. Perturbations are visible for human eyes. We present the attribution error as well as SSIM. }
    \label{fig:adv_example}
\end{figure}


\noindent{\bf 2. Adversarial Perturbations in the Image Space:}  
This class of adversarial perturbation instead considers the feature extractor only. Our intuition is to make it easier to optimize for an effective adversarial perturbation if we only have to differentiate through the feature extractor rather than the generator combined with the feature extractor (as done in our attack in the seed space previously). This means that we first generate a deepfake using a generator, then we take this image and look for a perturbation that will maximize the distance between the adversarial image's feature representation and the original deepfake's feature representation. 

Here, we also first consider the FGSM to find such an adversarial perturbation for varying  $\ell_\infty$ constraints $\varepsilon$. 
Figure~\ref{fig:error_v_eps_image} suggests that our method is robust under small values of $\varepsilon$, but will eventually deteriorate to random guessing as $\varepsilon$ increases. However, it is worth mentioning that because the $\ell_{\infty}$ norm only bounds the maximum change per pixel, artifacts introduced by the perturbation are more easily noticed by a human observer. We show the same deepfake perturbed with $\varepsilon=0.0588$ and $0.1$ in Figures~\ref{fig:adv_015_target_0} and~\ref{fig:adv_01_taraget_0} respectively. Perturbations of this size are discernable by humans, especially on the background region for each image. As seen in Figure \ref{fig:adv_example} (d), the background of it has noticeable artifacts that can be used to to identify as an adversarial example. We also calculate the structural similarity (SSIM) between the perturbed deepfake and the original deepfake to evaluate human perception more accurately; as $\varepsilon$ increases, the SSIM decreases.

We find that the $\ell_2$ variant of the Fast Gradient Method is unsuccessful in this setting as shown in Figure~\ref{fig:error_v_eps_image}, possibly due to the formulation our loss for this attack.
For this reason, we turn the formulation by Carlini and Wagner (CW)~\cite{carliniwagner} and adapt it to our problem. We have two goals---minimize the perturbation under the $\ell_2$ norm and maximize the feature difference. We use $c$ to weight the two goals, as shown in the following equation:
\begin{equation}
    \operatorname*{argmin}_\delta \norm{\delta}_2 - c \cdot d({\bf{x}}, {\bf{x}}+\delta) 
\end{equation}
In Figure~\ref{fig:error_v_c_carlini}, we highlight the results we obtain using perturbations generated with the CW-based approach. As before, the attribution process is robust to a small perturbation (\ie small values of $||\delta||_2)$ and degrades as the perturbation size increases.


We formulate a few hypotheses to explain why crafting small adversarial perturbations against this attribution process could be more challenging than directly attacking a ML classifier. First, generator models map a low-dimensional seed space to a high-dimensional image space. This changes the nature of the optimization problem solved to find adversarial perturbations. Second, the distance metric we use to compare two images uses an $\ell_2$ norm which averages out some of the perturbations introduced and decreases the impact on the attribution's error rate.  

\noindent{\bf 3. Transferability attack on black-box attribution:} To contrast our efforts with prior work~\cite{frank2020leveraging}, we design a simple experiment which treats the attribution process as a black-box classification model. We then craft adversarial examples that evade the classifier and test how likely they are to evade not only the classifier, but also our attribution mechanism from \S~\ref{sec:reconstruction}. In other words, we test how transferable adversarial examples crafted on substitute models are to our black-box attribution process. Our intuition here is to show that it is easier to find informative gradients on the classifier than on the attribution process itself which involves a non-differentiable optimizer. This helps us more rigorously evaluate the worst-case performance of our attribution process in the presence of a motivated adversary.

We train a substitute classifier on 24,000 synthetic images (8,000 from each GAN); the classifier comprises of 4 convolutional layers, and its objective is to classify each image to one of the 3 GANs (ideally the one that created it). The classifier reached over 99.8\% accuracy on a test set composed of 3,000 images. 
We then attacked the classifier using FGSM and CW perturbations respectively. 

\begin{figure}[t]%
    \centering
    \includegraphics[scale=0.35]{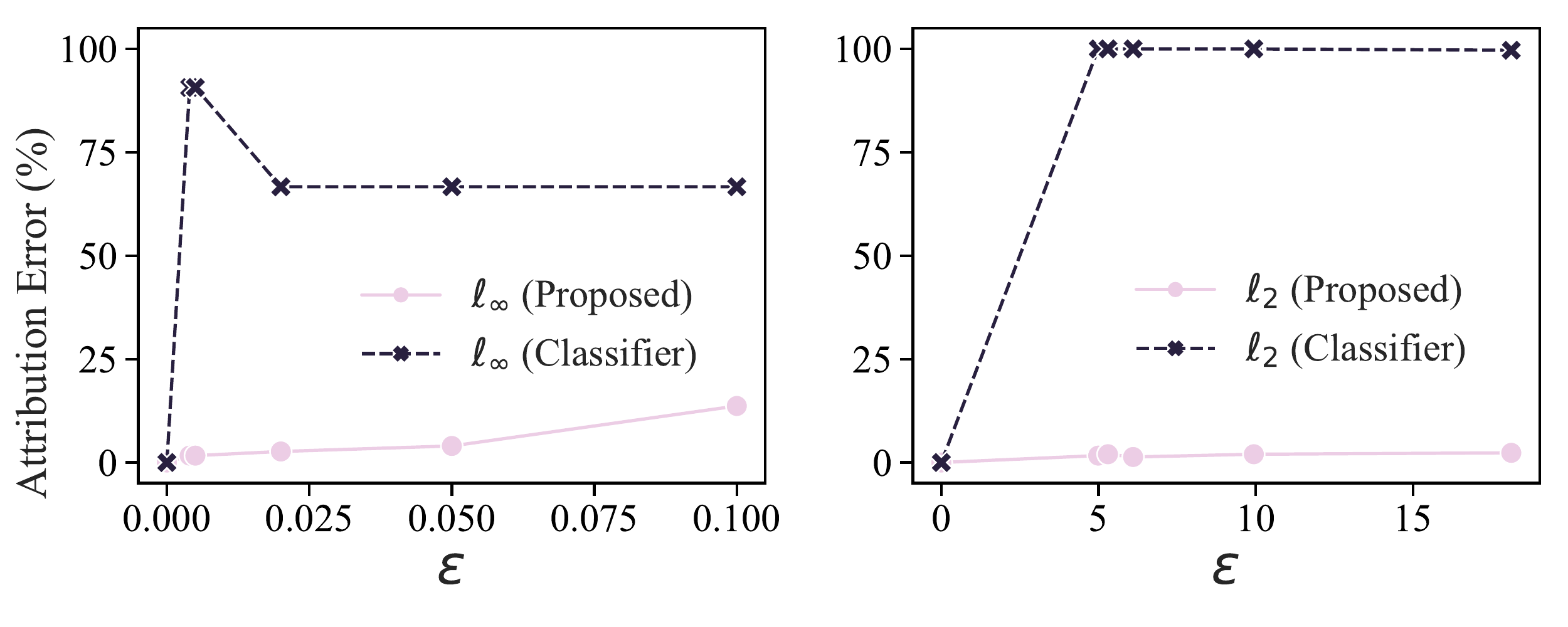}
    \caption{Robustness of the classifier and our attribution algorithm to adversarial examples crafted on the classifier. Adversarial examples are generated with the FGSM and  CW attacks, constrained using $\ell_\infty$ norm (top) and $\ell_2$ norm (bottom).}
    \label{fig:adv_transfer}
\end{figure}

Results for both $\ell_\infty$ and $\ell_2$ perturbations are shown in Figure~\ref{fig:adv_transfer}. The classifier is extremely sensitive to adversarial perturbations; the CW perturbation is particularly effective and is able to reduce the classification accuracy to 0\%. On the other hand, when we transfer these adversarial examples to our attribution method they are ineffective. This means that transferability-based attacks that use small perturbations (as discussed earlier, when the perturbation is too large, attribution fails as expected) fail to transfer. The attribution accuracy only drops by 1.7\% in the case of CW perturbation, and less than 15\% in the extreme case of FGSM perturbation (for large values of $\varepsilon$ in the $\ell_{\infty}$ regime).

\subsection{Fine-Tuned Models}
\label{sec:fine_tune}
Previous sections demonstrated that  relaxed attribution is effective in various benign and adversarial settings. Here, we consider relaxed deepfake attribution in a more difficult setting -- a setting that tries to attribute deepfakes to generators that are almost exact replicas of each other. We seek to attribute images generated by a set of several fine-tuned models i.e. related parent-child models within a number of optimisation steps between them. Successful attribution in such a setting suggests that relaxed attribution is possible even when models are extremely similar, further highlighting scalability and robustness of the method. 

We fine-tuned author-provided ProgressiveGAN, StyleGAN and StyleGAN2 with additional twenty-, hundred- and five hundred- thousand images.  Do note that the fine-tuning step here is minimal -- ProgressiveGAN is trained with 12 million images, whereas StyleGAN and StyleGAN2 are trained with 25 million images. We used standard training parameters from each of the models respectively. We chose those parameters, as they represent one, five and twenty five units of training steps in the original respective GAN implementations. We fine-tuned ProgressiveGAN, StyleGAN and StyleGAN2 with the steps outlined above to get 9 children models. We combined them with the original parent models to get an overall of 12 models. We then ran relaxed attribution method with three reconstruction attempts. Do note that setup here is significantly harder than the one described in the previous section -- there are more models involved, all models are related, and there is only a small difference between them.  Despite increased complexity of attributing very similar, related models, we find that out method achieved 90.9\% (1091/1200) accuracy, compared to 97.62\% accuracy in a non-malicious setting (\S~\ref{sec:benign})\footnote{Interestingly, we find that when the true originating model is not present in the model pool, attribution often hints to its closest relative in the pool. See \Cref{sec:finetune_notrue} for more details.}.

\subsection{User Study}
\label{sec:user_study}

\begin{table}[t]
    \centering
    \resizebox{\columnwidth}{!}{
    \begin{tabular}{l|l|c|c|}
\multicolumn{2}{c}{}&\multicolumn{2}{c}{Participant Attribution}\\
\cline{3-4}
\multicolumn{2}{c|}{}&Incorrect&Correct\\
\cline{2-4}
\multirow{2}{*}{Model Attribution}& Incorrect & $2$ & $0$ \\
\cline{2-4}
& Correct & $37$ & $592$ \\
\cline{2-4}
\end{tabular}
}
    \caption{Confusion matrix for participant and model attribution results. There is a strong correlation between participant and automated machine learning model attribution decisions although participants performed slightly worse than the model.}
    \label{tab:attribution_cm}
\end{table}


In this section we investigate the extend to which solutions found by our method are interpretable by humans. It is particularly interesting to compare humans performance in measuring distances between the different reconstructed images produced by our seed reconstruction algorithm and the original deepfake image that the reconstruction targets. 

\begin{figure}[t]
    \centering
    \fbox{\includegraphics[scale=0.2]{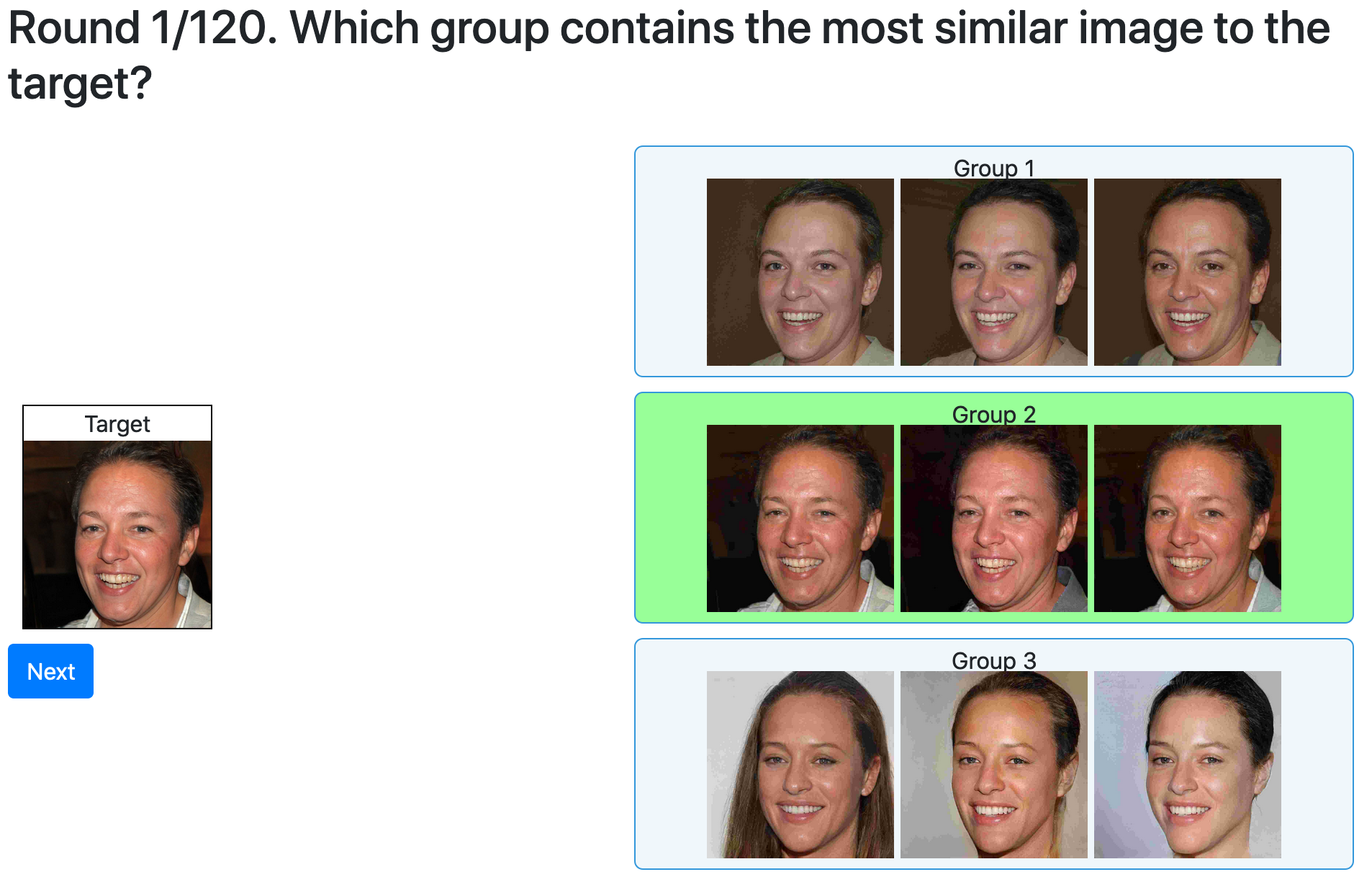}}
    \caption{Screenshot of one round of the user study. Each round randomly selects a target image and shuffles group order.}
    \label{fig:user_study}
\end{figure}

\noindent{\bf Experimental Setup:} We carried out a user study on Amazon Mechanical Turk, where we recruited 122 participants to identify which of the reconstructed images (using 3 random seeds, and 3 GANs) is visually most similar to the synthetic target. Note that the synthetic image (\ie the target) is selected randomly, and the order of the reconstructed images are shuffled for each round. An example task provided to the participants can be found in Figure~\ref{fig:user_study}. As the figure illustrates, reconstructed images from the same GAN are composed into one row and each participant was asked to complete 120 rounds. This study was approved by our Institutional Review Board, and each participant was compensated 2 USD for their efforts. The user study interface is designed with OTree~\cite{CHEN201688}.

To ensure the quality of data from participants, each round has a probability of 5\% being a placebo round where one row has three exact copies of the target. We calculated a participant's average placebo round accuracy and discarded the data from participants that achieved less than or equal to 80\%, leaving us 111 valid participants in total. The results and analysis discussed below are pertinent to the valid participants only and we simply refer to them as participants. 
In total, our data collection led to 3246 synthetic images receiving 5710 votes from the participants, with an average of 1.76 votes per image and standard deviation of 1.03. In order to investigate the correlation of attribution between the algorithm and the participants, we select 631 synthetic targets that received at least three votes from all participants (regardless if the attribution is correct or not).

\noindent{\bf Results:} On average, across all images shown, participants achieved 89.81\% attribution accuracy with standard deviation of 5.84\%. In comparison, our approach (relaxed attribution) achieves 97.6\% attribution accuracy (refer \S~\ref{sec:benign}). On 631 images that have received at least 3 votes, we calculated the majority votes for each with participant-algorithm matching matrix shown in Table~\ref{tab:attribution_cm}. Out of 631 images, human and our method agree 93.8\% of the time. This indicate a strong association between human judgement and the decision metric we use. Noticeably, no single image has been attributed incorrectly by the algorithm but correctly by the participants at the same time. We believe our algorithm can be used collaboratively with human forensic experts to provide a higher attribution accuracy. This also paves the way to provide explainability for the attribution algorithm, as decisions are made through visual reasoning. We leave a more rigorous explainability analysis to future work.

\section{Related Work}
\label{sec:related}

Methods for detecting if multimedia is authentic or synthesized can broadly be classified into two categories:
\begin{enumerate}
    \item The first is based on conventional image processing techniques; synthesized (or altered) images do not follow the same physical principles that authentic (or real) images capture. For example, photos that are captured through a camera's optical system will result in distortions such as misalignments in color channels. Identifying such inconsistencies paved the way for digital forensics~\cite{8272035}. 
    \item The second category involves comparing noise levels in certain areas of images to that of the whole image~\cite{popescu2004statistical}. Using deep learning techniques, one can extract features and learn a binary classifier to distinguish between authentic and synthesized images. The work of Wang \etal~\cite{wang2019cnn} achieves high performance in distinguishing real images from those synthesized by GANs despite being trained on synthesized inputs from ProgressiveGAN~\cite{karras2017progressive} alone. A  detailed survey of the second category of approaches is found in the work of Verdoliva \etal~\cite{verdoliva2020media}.
\end{enumerate}
As detailed in the introduction, we seek to attribute deepfakes rather than detect them. We explained how this is mainly because detecting deepfakes is only likely to fuel advances in the generation of more realistic deepfakes that become increasingly harder to detect. Instead, we focus on attribution and we seek to assist forensics experts in their work.

Digital forensics is essential for many applications; in social media platforms, providing evidence if media is synthesized draws a clear distinction between moderation and censorship. For example, Twitter has been labeling tweets that contain synthesized content since February 2020~\cite{twitter_2020}. Additionally, such an identification procedure also aligns with current regulations in requiring explanation for machine-made decisions, such as the GDPR framework in EU.

One way to attribute a synthesized image to the model that generated it is to identify fingerprints that exist in synthesized images~\cite{yu2018attributing,frank2020leveraging}. However, such fingerprinting techniques have been shown to not robust against small image perturbations, or adversarial attacks~\cite{carlini2020evading}. 

\section{Discussion}

\subsection{Limits of attribution}
\label{sec:problem_def:limits}

There are practical limitations to attribution which we identified through our evaluation in \S~\ref{sec:eval}. These limitations arise from cases where GANs are similar, as outlined in our experiments on fine-tuned models in~\Cref{sec:fine_tune}. One can see how two quasi-identical GANs with an insignificant difference in a single weight would learn the same or near identical biases and make attribution difficult. Here, we provide an analytical argument as to why attribution with strict integrity---per the definition in \S~\ref{sec:problem_def}---is not always possible. This observation holds even in the setting where  GANs converge to different synthetic distributions upon completion of training. Our analytical example constructs a GAN architecture such that any arbitrary synthetic image can be attributed to it. This would mean that an adversary could have any synthetic image misattributed to the developer of this GAN architecture.

We generalize the observation of Abdal \textit{et al.}, which showed that one can configure StyleGAN in a way that it produces an arbitrary image~\cite{abdal2019image2stylegan}, to draw the conclusion that other types of GANs can also be fed seeds that lead them to generate an arbitrary image.
The reason why this is possible lies in one of the secondary goals of GAN training. A generator model should not only produce realistic synthetic samples which could have been drawn from the data distribution $p_{data}$, it should also produce diverse synthetic samples. Otherwise, the generator will exhibit failures such as mode collapse~\cite{2016arXiv160603498S}: it will default to generating synthetic samples that are close to training images it learned from. In addition to the seed required at the input of the generator, many modern generator architectures promote diversity by injecting randomness into all layers of the network, often up to the last layer. Here, we show how this randomness allows the generator to synthesize diverse content, with the unfortunate consequence that an adversary could, in the limit, force generators to synthesize an arbitrary image. In this setting, attribution with strict integrity is not possible. 


As illustrated in StyleGAN's implementation~\cite{stylegan_github}\footnote{We include a copy of StyleGAN's structure in \Cref{fig:stylegan_key} in \S~\Cref{sec:sup_figures}.}, the first architectural element promoting diversity is the \textit{style} parameter $\mathbf{A=\{A_\gamma, A_\beta\}}$, which is derived by applying fully-connected layers to the seed $\mathbf{s}$. For instance, in our experiments on face generation, the style parameter $\mathbf{A}$ is used by StyleGAN to exert control over the face styles, including the presence of glasses, age, or the face's orientation. A second source of entropy is provided to each   layer to further promote diversity. Each layer's input $\bf{x}_i$ is combined with a weighted random input $\mathbf{B}$, before applying Adaptive Instance Normalization (AdaIn in \Cref{fig:stylegan_key}) to enforce style $\mathbf{A}$, per:



\begin{equation}
  \label{eq:stylegan}
  \begin{gathered}
    \overline{\mathbf{x}_{i}} = \text{ReLU}(\mathbf{x}_{i}\mathbf{+B \cdot w_{B}}) \\  
    \mathbf{x}_{i+1} =\text{AdaIN}(\overline{\mathbf{x}_{i}}, \mathbf{A}) = \mathbf{A}_{\gamma}\dfrac{\overline{\mathbf{x}_{i}} - \mu(\overline{\mathbf{x}_{i}})}{\sigma(\overline{\mathbf{x}_{i}})} + \mathbf{A}_{\beta}
  \end{gathered}
\end{equation}




If one is able to control either the source of randomness $\mathbf{B}$ or the style $\mathbf{A}$ being fed to the last layer, the model can be forced to output any arbitrary sample. To see why, observe how an attacker could set $\mathbf{A}_{\gamma,i}=0$ and $\mathbf{A}_{\beta,i}$ to a target image. This will result in the generator synthesizing the target image. The same effect can be achieved by carefully setting the value of $\mathbf{B}$: \Cref{eq:stylegan} can be inverted to produce an image of the adversary's choice. 


\subsection{Plausible deniability}
\label{sec:problem_def:pd}

In light of this discussion, we turn back to our motivation for attribution. The logical conclusion is that strict integrity (see \S~\ref{sec:problem_def:limits}) for the attribution of a deepfake cannot be solely obtained by analyzing the manifold learned by each generator. In \S~\ref{sec:relaxed_integrity} and \S~\ref{sec:eval}, we introduced and evaluated an approach that is nevertheless able to achieve relaxed integrity. 
The lack of strict integrity guarantees could however lead to GAN developers refraining from sharing their GANs to prevent misuse and avoid any repercussions.
For example, MIT was recently pressured to remove an dataset over the abusive data and labels inside of it~\cite{mittinyimage}. A recent challenge ran by Facebook illuminated the significant complexity of deepfake detection~\cite{deepfakefbchallange}, whereas in this paper we describe why detection is deemed to fail in the long term. 
Thus, we propose that in settings where strict integrity is needed, we move away from detection and attribution to instead provide plausible deniability. 

Despite apparent problems with plausible deniability in the real world such as with Anti-Money Laundering and Know Your Customer regulations~\cite{anderson2019bitcoin}, it does help to identify a responsible party and stop outright abusive behaviour. We argue that similar legislation is needed to frame the deployment of ML systems. 
In particular, plausible deniability helps tilt the asymmetric relationship between legitimate and malicious uses of generators in favor of legitimate users. Currently, an attacker can use modern anonymity-preserving technology to hide their identity, whereas it is harder to anonymize the generation process and the generated data itself. We believe that, through attribution, methodology described in the paper contributes to transparency and auditability around the use of generative modeling. This naturally allows for an ethical user of generative technology to help in a responsible manner: if developers of generative technology have responsibilities to support the ability to provide plausible deniability, then this could make malicious behaviour less easy to engage in. This would overall improve the status quo around deepfakes. 

In attribution, we rely on a trusted third party to attribute the synthetic example to a generator. This could be for instance a law enforcement agency investigating the use of a deepfake. The law enforcement agency has a set of suspected generators and performs attribution with or without participation from the different model developers. This detailed forensic analysis could for instance help `traitor tracing`. Instead, when it comes to \textit{plausible deniability}, the model developer is responsible for providing evidence that they did not generate the deepfake. This class of approaches will alleviate some of the possible repercussions faced by model developers when they release generative models, so long as all legitimate uses of the model are willingly and transparently monitored. If this level of accountability is put in place, this makes it easier to provide strict integrity---per the definition in \S~\ref{sec:problem_def}---because the model developer can easily capture the state of random inputs in addition to the synthetic example itself. We discuss such an approach in~\Cref{sec:strict_integrity}, where we construct a hyperledger that records everything that a model ever produced. This allows for provenance of content synthesised by GANs in a distributed fashion and enables a forensics expert to confirm that a given deepfake was not produced by a given model.

\section{Conclusion}

In this paper we have looked into attribution and plausible deniability for deployment of synthetic content generators. 
We show analytically and practically that as generation becomes better, attribution becomes harder, up to a point where a sample can be attributed to all of the models and is indistinguishable from real data. Ultimately, this hints that attribution is not just a computer science problem, but requires a solution incorporating operational security, legal and ethical frameworks. Furthermore, given the fragility of ML to adversarial examples, it is imperative that humans should be able to inspect and interpret the decision of a ML model. 

In this paper we  presented one possible solution. We showed that a synthesized image can be attributed to the generator that produced it even in presence of noise. Further, we find that the attribution performance is supported by a human's decision in 93\% of the cases. Of the failed cases, we find that more than half were because the method failed to reconstruct the images even after three attempts. Yet, it is clear that for attribution to be successful in the future, ML tools are not enough. With fake images indistinguishable from real data, information hiding, fingerprinting, and watermarking become a necessity for both data sources~\eg see Shafahi et al.~\cite{shafahi2018poisonfrogs} and models~\eg see Shumailov et al.~\cite{shumailov2019sitatapatra}.

\bibliographystyle{plain}
\bibliography{usenix}

\newpage

\appendix
\section{MNIST Evaluation}\label{sec:mnist}
To validate the efficacy of the relaxed attribution method, we also evaluated it on two generative models trained with MNIST data to produce fake handwritten digits. This was to ensure that our results are not specific to the face generation task considered in the main body of the paper.

\subsection{Implementation}

MNIST, which is a handwritten digits dataset that contains 60,000 training and 10,000 testing images, was widely adopted in the initial developments of 
GANs. We selected two models compatible with MNIST: the \textit{DCGAN}~\cite{radford2015unsupervised}~\cite{dcgan_tensorflow} model contains 3 transpose convolution layers in its generator; the \textit{Brownlee}~\cite{brownlee_gan} model contains 2 transpose convolution layers in its generator, with a smaller filter size in its generator. Input seeds to both models are 100-length float vectors. They also have similar discriminators except the one in \textit{DCGAN} model has a slightly large filter size.
We trained both models with Adam optimizer and 0.1 as learning rate for 50 epochs. 

\subsection{Evaluation}

\paragraph{Generate Dataset} To evaluate our method on MNIST-based synthesizers, we generated 1000 synthesized images with random seeds, with 500 on each model. 

\paragraph{Classification Method} Due to the simple nature of MNIST images, we used the $\ell_2$ difference between reconstructed images and target images as the loss function for reconstruction. For each target image in the 1000 image dataset (500 for each model), we select 1 random seed to reconstruct the target image on both models. Each reconstruction is done with 500 optimization steps of Adam as the optimizer. 

\paragraph{Results} For each target image, we select the model that produces the smaller image difference during reconstruction as the identified model. Our method achieved 94.6\% classification accuracy, with attribution results shown in Table~\ref{table:mnist_result}.

\begin{table}[t]
    \caption{MNIST attribution results.}
    \label{table:mnist_result}
  \centering
  \begin{tabular}{lll}
    \toprule
    \cmidrule(r){1-2}
    &  DCGAN  & Brownlee Model \\
    \midrule
    DCGAN-generated images & 498 & 2 \\ 
    Brownlee-generated images & 49 & 451 \\ 
    \bottomrule
  \end{tabular}
\end{table}



\paragraph{Additional Observations}
\begin{enumerate}
    \item By observing loss curves, we noticed that when reconstructing a target image on the model that generated it, our method can always result in near zero reconstruction image loss. 
    \item However, original seeds are unlikely to be recovered. Even if the initial reconstruction seed is set to be the original seed plus a small amount of noise $\mathcal{N}(0,\,0.0025)$, none of the original seeds were recovered in 1000 experiments. This means that latent space has a large number of collisions.
    \item Additionally, we observe that most failed attribution cases happen when the Brownlee model fails to reconstruct the target, but DCGAN produces a reasonably good reconstruction. We speculate that  DCAGN has an advantage in reconstruction power compared to the Brownlee model since it has more layers. 
\end{enumerate}

\section{Errors in seed reconstruction}
\label{sec:seed_difference}
To evaluate fidelity of seed reconstruction, we measured the $\ell_2$ difference between the initial seed ${\bf{s}}_i$ and the reconstructed seed ${\bf{s}}_e$ after optimization completes. Out of 18,000 reconstructions on 6,000 images with 3 seeds per synthesized image, none of the reconstruction attempts resulted in a smaller $\ell_2$ difference between the reconstructed seed and the seed that was originally used to synthesize the image upon completion of the optimization procedure---despite the corresponding synthetic image being close to the target deepfakes. 
The average initial difference between the candidate seed and the original seed is 1.99, whereas the average difference after the reconstruction completes is 13.00. We find that seeds after optimization tend to take larger values, compared to random seeds which are typically sampled from a Gaussian distributions. This confirms that the goal for the relaxed attribution algorithm is not to recover the original seed, but to find collision seeds that produce similar synthetic outputs. Follow-up work may also find that additional constraints to the problem in \Cref{eq:reconstruction} improve the similarity between reconstructed and original seeds.

\section{Attribution among fine tuned models when ground truth does not exist}
\label{sec:finetune_notrue}
In this evaluation, we attempt to attribute parent-child set of models without the correct model present. We repeated the same experiments as is described in~\Cref{sec:fine_tune}, but only with 11 models present -- we exclude the original generating model for each image during attribution. Here, we assume attribution to be correct, if the sample can be correctly attributed to the originating model architecture. A random guess in this case will result in an accuracy of 3/11 or 27.3\%. In our experiments, the relaxed attribution algorithm achieved 98\% (1176/1200) accuracy with reconstruction sample size of 3. The high success rate demonstrates that the relaxed attribution algorithm can successfully attribute a given sample to a model with same architecture (in our case fine tuned checkpoints) when the exact model is not present in the pool.

\section{Providing Plausible Deniability within Strict Attribution}
\label{sec:strict_integrity}
\subsection{Plausible Deniability May be Achieved with a Non-ML solution}
While it may be possible to use a ML-based solution to attribute deepfakes in the \textit{relaxed integrity} setting, we believe that \textit{strict integrity} requires a non-ML solution to avoid an arms race between the development of attribution mechanisms and attacks against these mechanisms. It is theoretically possible that, given an image, one can find a seed that will result in a generative model generating that image almost perfectly. We discussed this in \S~\ref{sec:problem_def:limits}. Thus, a malicious entity is likely to be able to forge a seed which would result in a deepfake being incorrectly attributed to an innocent model developer. 

In this section, we instead turn to traditional techniques from computer security literature to demonstrate how developers of generative models can prove that they did \textit{not} generate a deepfake. Our goal is thus to achieve \textit{plausible deniability}. We discuss our proposed solution with the potential of legal action being taken against model developers in mind.

\subsection{Blockchain-based Attribution Solution}
Inspired by how blockchain is used to provide data provenance in cloud computing environment~\cite{7973733}, where data can be publicly audited and tamper-proof, we propose a blockchain-based solution to store image generation record to provide plausible deniability for image generators. The proposed solution contains two processes: image generation and image validation. Following are the critical components of the proposed solution: 

\begin{itemize}
  \item \textbf{Image Generator} An image generator, denoted as G, is any entity that produces a generated image and wants to opt for the proposed solution to obtain plausible deniability. The identity of an image generator is provided by a public-private key pair, denoted as $k_B$ and $k_P$ respectively. It is important to note that although an image generator has access to generation models, it can not make any IP claim with this system.
  \item \textbf{Model Database} A model database is the place to store generation models submitted by image generators. The database needs to be publicly accessible for read, but can be either centralized or decentralized based on image generators' preference. Different image generators can maintain independent model databases if preferred. It is in the best interest of image generators to ensure reliable access to their generation models via model databases when requested by the system. 
  \item \textbf{Blockchain} The blockchain network consists of globally participating nodes. All image generation records will be stored as block entries and verified by blockchain nodes.  
\end{itemize}

\begin{figure}[t]
  \centering
  \includegraphics[width=1\linewidth]{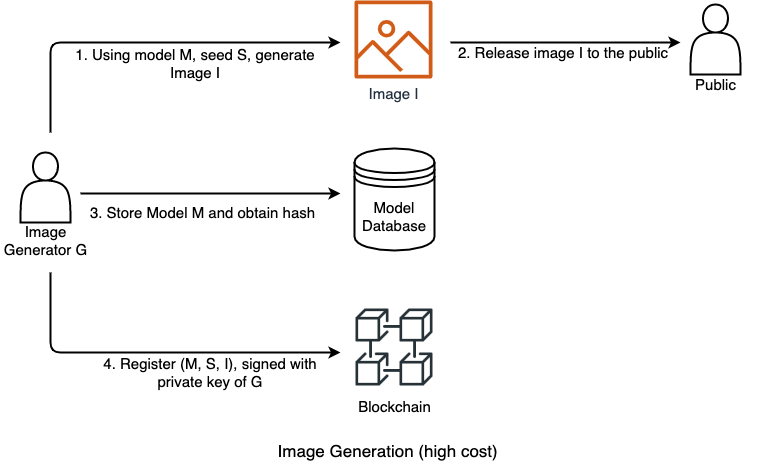}
  \caption{High-level diagram of image generation process with \textit{strict integrity attribution} solution.}
  \label{fig:blockchain_generation}
\end{figure}

\begin{figure}[t]
  \centering
  \includegraphics[width=0.7\linewidth]{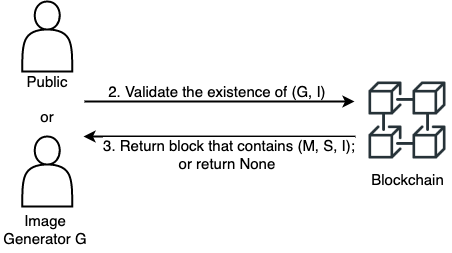}
  \caption{High-level diagram of image validation process with \textit{strict integrity attribution} solution.}
  \label{fig:blockchain_validation}
\end{figure}

\subsubsection{Preliminary Concepts}
\noindent{\bf Structure Definition of a Blockchain Entry:} As shown in Table~\ref{table:entries}, each entry contains a record ID, time of generation, hash of generation model, seed used (with padding), hash of generated image and is signed by the image generator with its private key. \\

\begin{table}[t]
  \centering
  \resizebox{\columnwidth}{!}{
  \begin{tabular}{cccccc}
    \toprule
    \cmidrule(r){1-2}
    {\bf RecordID}     & DateTime   & ModelHash(a-bits) &  Seed(b-bits) &  ImageHash(c-bits) & Signature(d-bits) \\
    \midrule
    \midrule
    001&2021-01-20 12:00:00&Hash(Progressive)&Seed1+padding&Hash(Image1)&Hash(Alice)\\
    002&2021-01-20 13:00:00&Hash(Stylegan)&Seed2+padding&Hash(Image2)&Hash(Bob)\\
    003&2021-01-20 14:00:00&Hash(Stylegan2)&Seed3+padding&Hash(Image3)&Hash(Charlie)\\
    \bottomrule
  \end{tabular}
  }
\caption{Example of blockchain entries}
\label{table:entries}
\end{table}

\noindent{\bf Node Verification:} When the image generator generator submits an request to the blockchain to append a new entry, a node on the blockchain system will start to verify whether the entry is valid. The node will first obtain all information provided within the entry, and additionally obtain the model from model database as instructed by the model generator. Then the blockchain node will run the model with the same seed as provided to generate an image on it's own system. If the hash of the newly generated image is the same as the image hash provided by the image generator, then the node will append the entry to the newest block. During this mining process, nodes that first finished the verification process will be rewarded by the system. 

\subsubsection{Working Processes}
Functionality of the proposed blockchain-based attribution system can be achieved with two main processes: image generation and image validation. 

\noindent{\bf Image Generation:} During image generation, an image generator G will choose a model M and a seed S to generate image I, as illustrated in Step 1 in Figure~\ref{fig:blockchain_generation}. The image generator can choose to release image I to the public. To register the image generator record with the attribution system, the image generator will then save the model M into the model database and obtain a hash of the model. If the model M already exists in the model database, the image generator can simply obtain the hash. Then, the image generator will submit a entry, which encapsulates the   information necessary to regenerate the image and the hash of the image, to the blockchain to be registered. Nodes on the blockchain network will verify the generation process and append the new entry to the chain. \\

\noindent{\bf Image Validation:} During image validation stage, the public or an image generator will try to validate whether an image I was previously generated by the image generator by submitting a query to the blockchain. The blockchain will respond with previously registered records if it finds any, or none otherwise.

\subsection{Practical Limitations}
There are a few limitations in implementing the proposed blockchain-based system in practice. First it is unclear whether current blockchain can support off-chain access to obtain the model from the model database, which is required when verifying the validity of new entries. This is refereed to as the oracle problem of blockchain. Second, to validate whether an image was indeed generated by a given model and seed, the node needs to rerun the generation process as part of the verification process. Since most generation processes involve heavy computations and complex operands, it is unclear whether current programming languages for smart contracts, like Solidity, will support this.

\section{Supplementary Figures}
\label{sec:sup_figures}

\begin{figure}[t]
  \centering
  \includegraphics[width=0.72\linewidth]{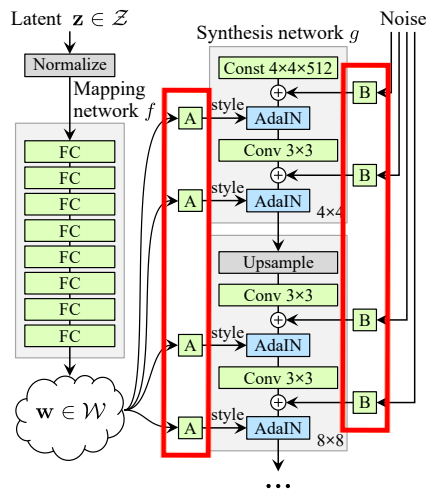}
  \caption{StyleGAN architecture with two generator input vectors: style $\bf{A}$ and randomness $\bf{B}$.  AdaIN is short for Adaptive Instance Normalization. Image adapted from~\protect\cite{karras2019style}. }
  \label{fig:stylegan_key}
\end{figure}

\section{Reconstruction Samples}\label{sec:recon_samples}
We plotted 4 synthetic targets and corresponding reconstruction bellow. The first three can be attributed successfully with our relaxed attribution method, while the last one failed due to diverged reconstruction on the correct model, StyleGAN in that case. Attribution results are bordered with green.

\begin{figure*}[t]
    \begin{minipage}{0.24\linewidth}
      \centering
      \includegraphics[width=\linewidth]{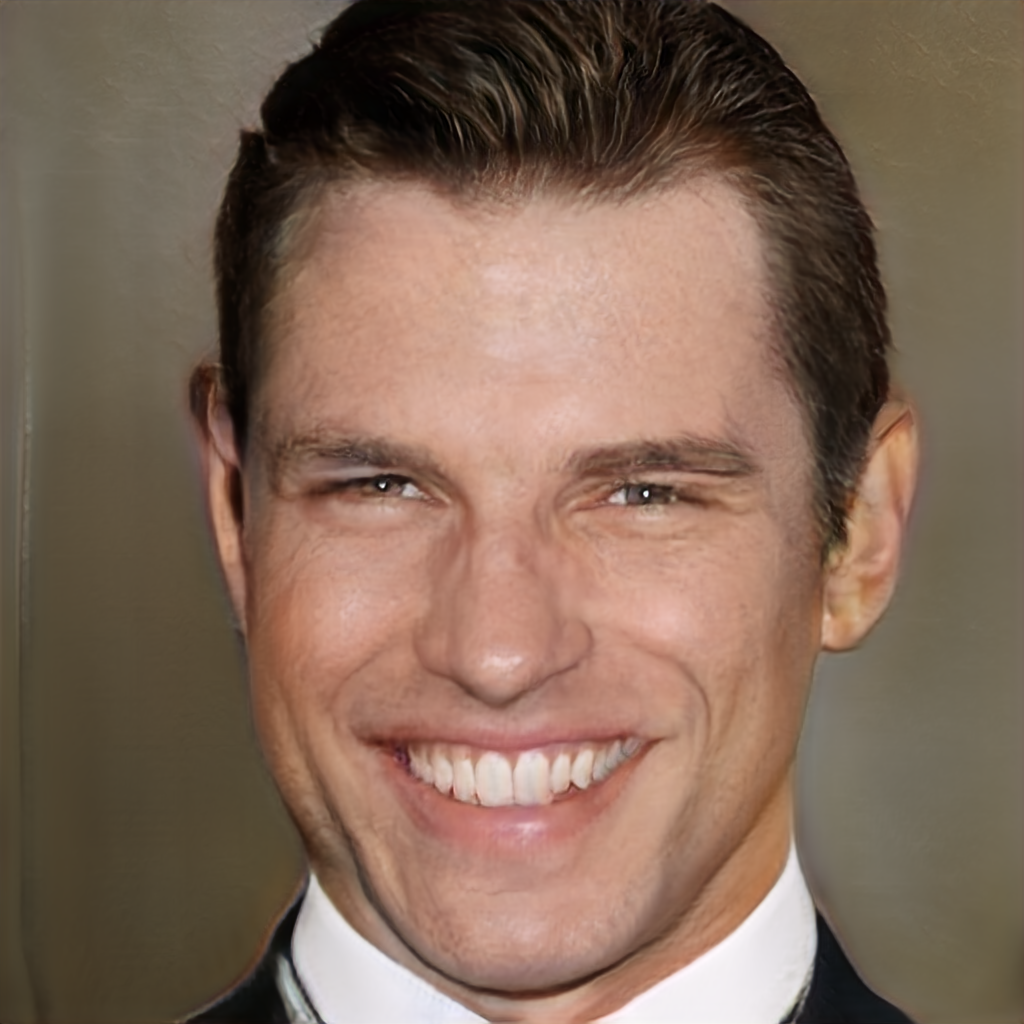}
      \caption{Target (from ProgressiveGAN)}
      \label{fig:faces_11}
    \end{minipage}
    \hfill
    \begin{minipage}{0.24\linewidth}
      \centering
      \includegraphics[width=\linewidth,cfbox=green 2pt 2pt]{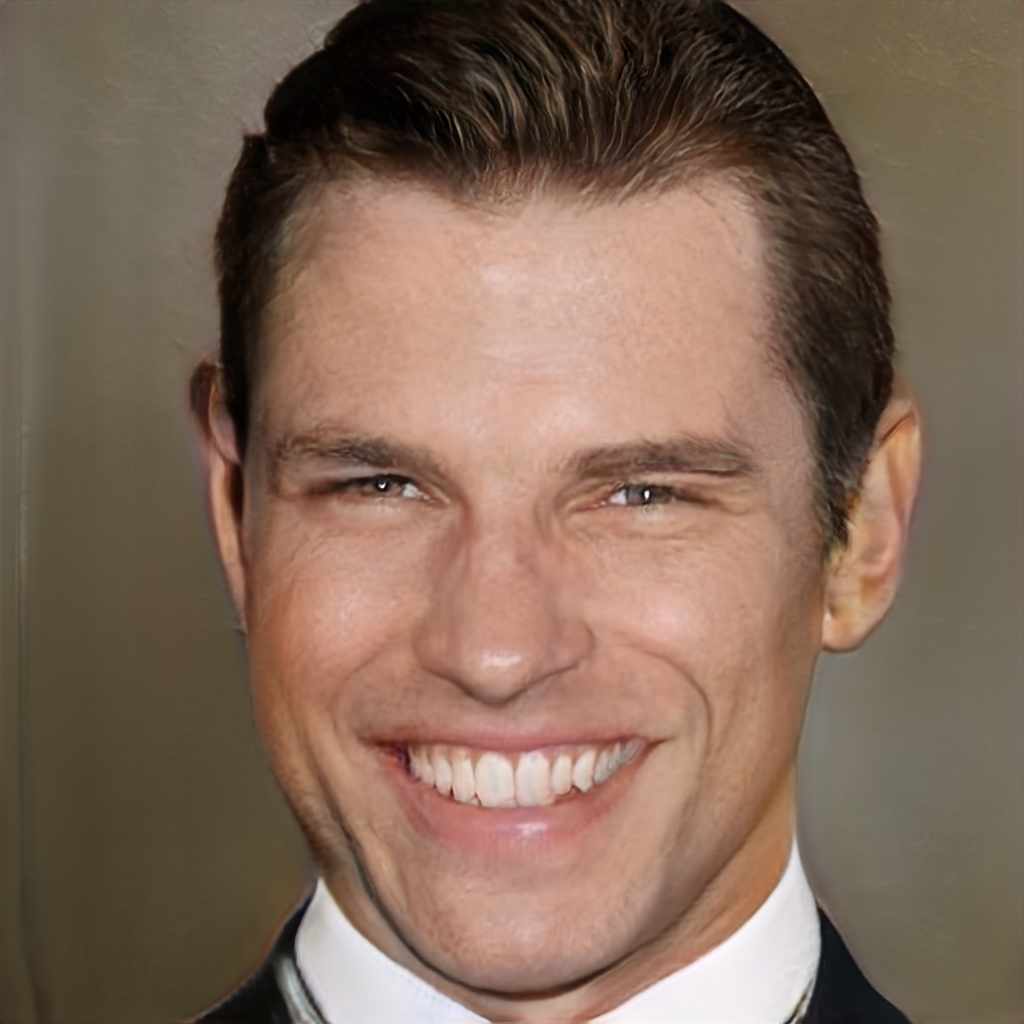}
      \caption{Reconstruction on ProgressiveGAN}
    \end{minipage}
    \hfill
    \begin{minipage}{0.24\linewidth}
      \centering
      \includegraphics[width=\linewidth]{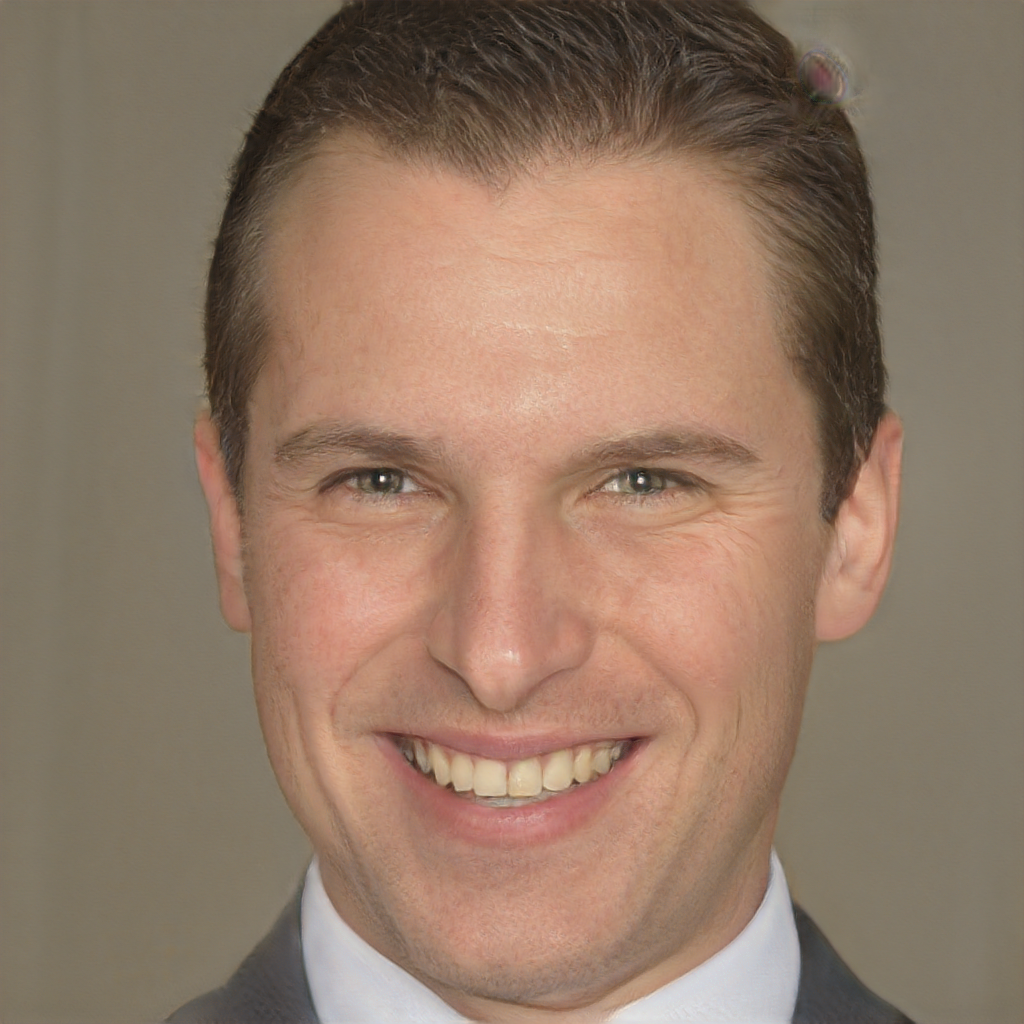}
      \caption{Reconstruction on StyleGAN}
    \end{minipage}
    \hfill
    \begin{minipage}{0.24\linewidth}
      \centering
      \includegraphics[width=\linewidth]{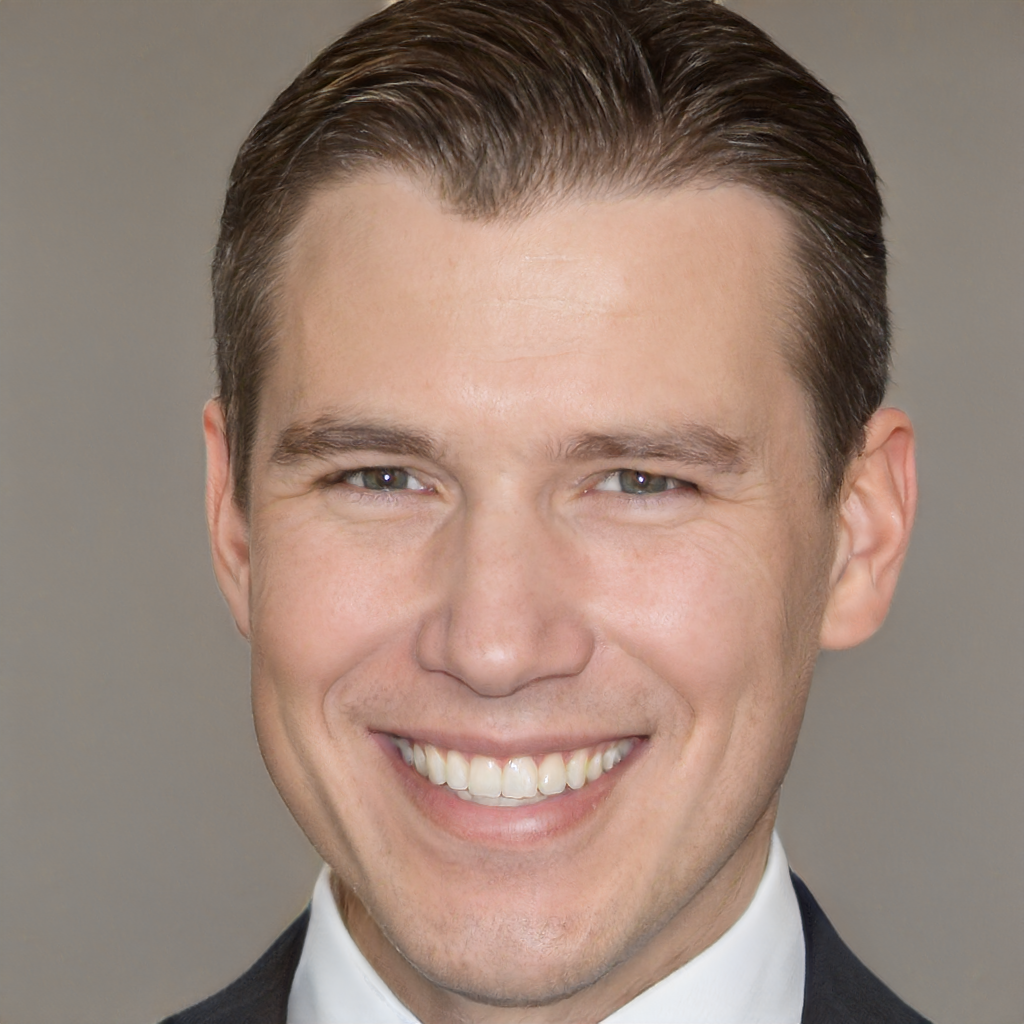}
      \caption{Reconstruction on StyleGAN2}
    \end{minipage}
    
    \begin{minipage}{0.24\linewidth}
      \centering
      \includegraphics[width=\linewidth]{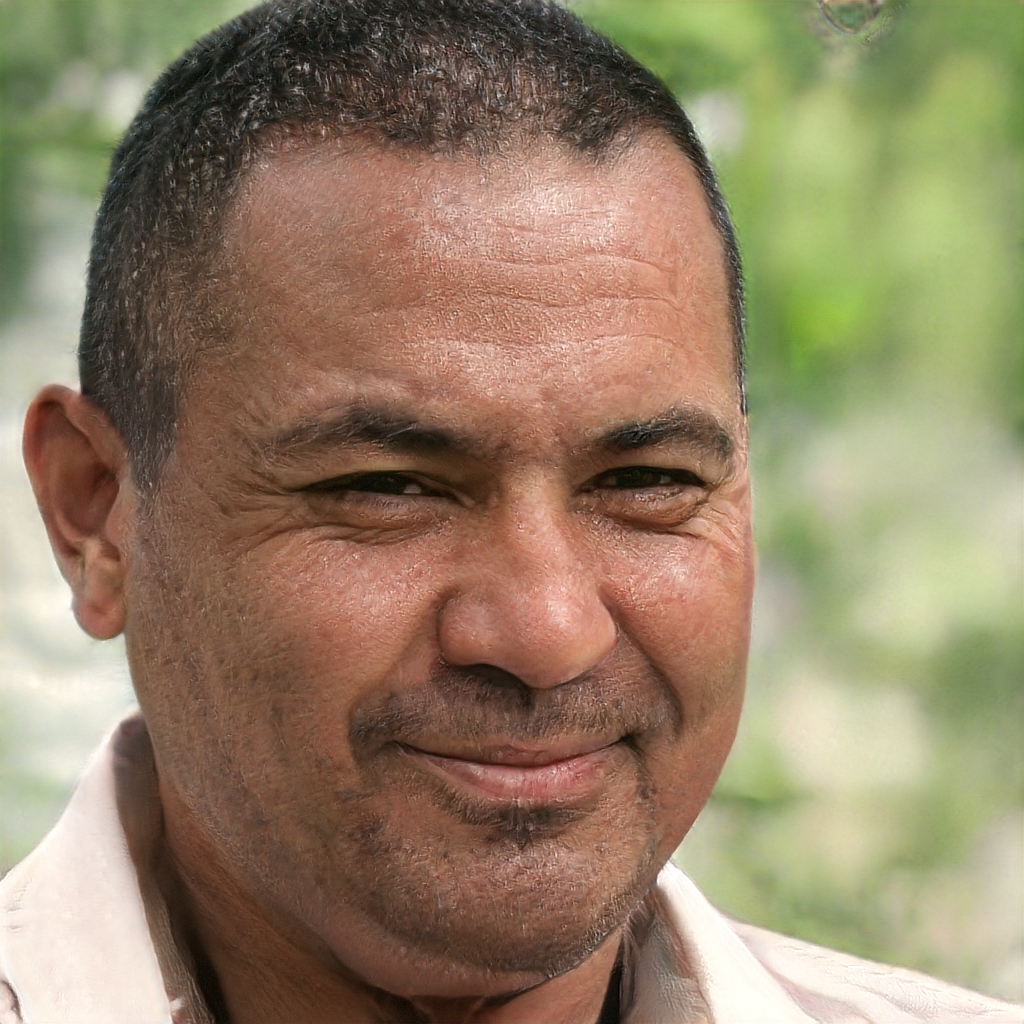}
      \caption{Target (from StyleGAN)}
    \end{minipage}
    \hfill
    \begin{minipage}{0.24\linewidth}
      \centering
      \includegraphics[width=\linewidth]{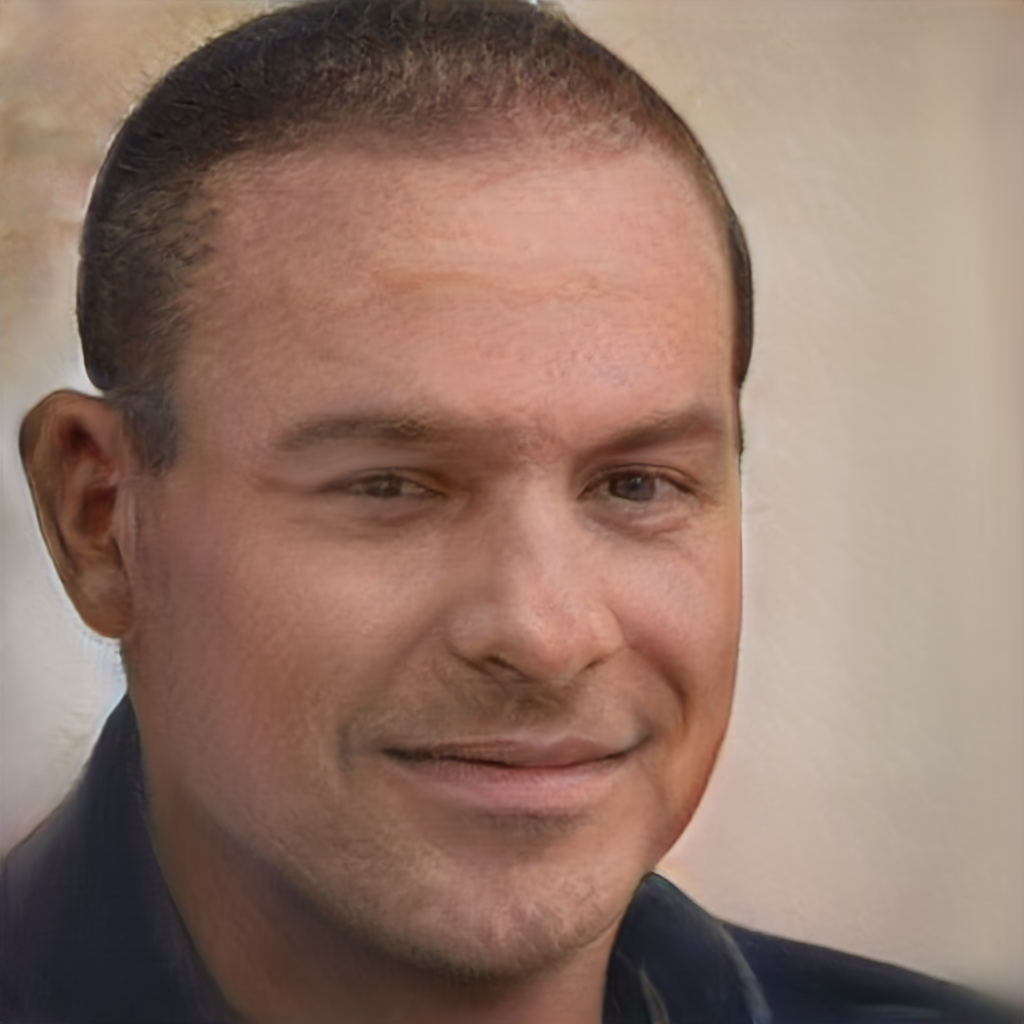}
      \caption{Reconstruction on ProgressiveGAN}
    \end{minipage}
    \hfill
    \begin{minipage}{0.24\linewidth}
      \centering
      \includegraphics[width=\linewidth,cfbox=green 2pt 2pt]{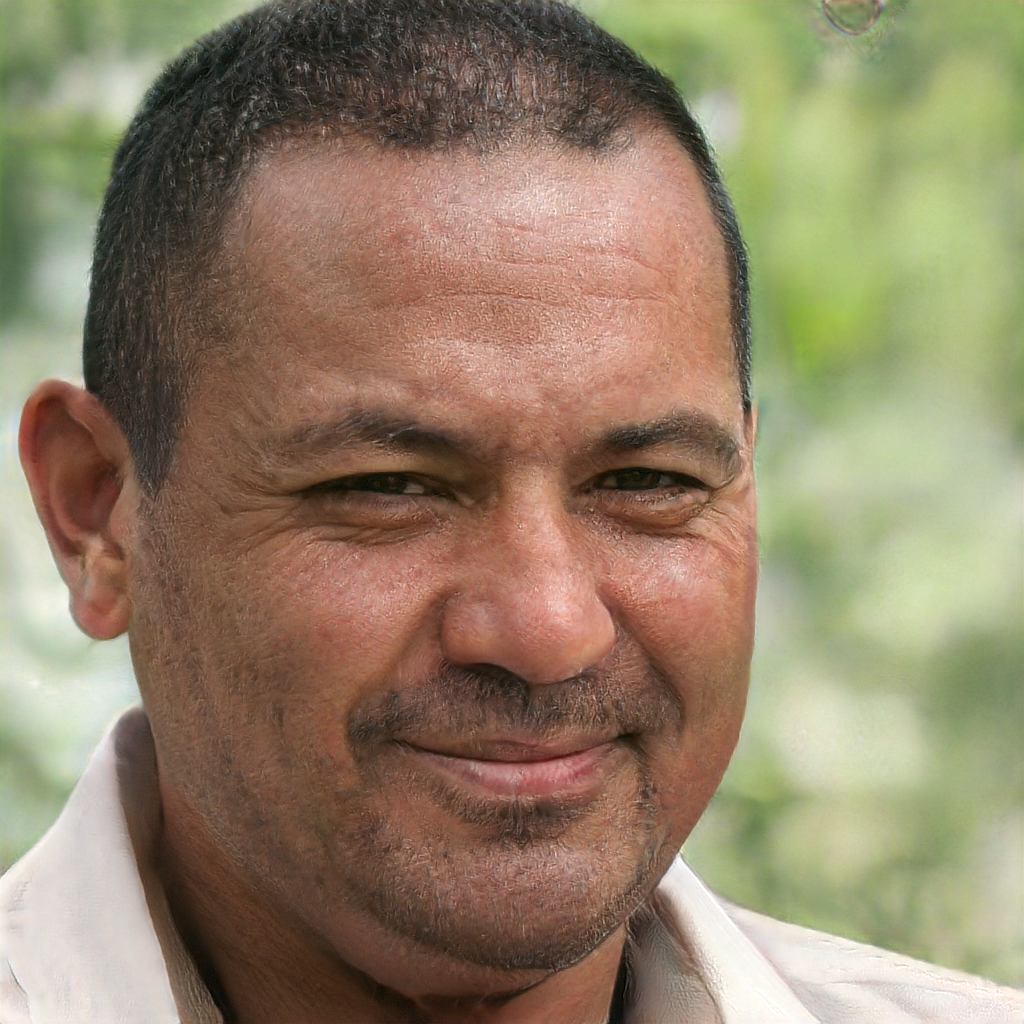}
      \caption{Reconstruction on StyleGAN}
    \end{minipage}
    \hfill
    \begin{minipage}{0.24\linewidth}
      \centering
      \includegraphics[width=\linewidth]{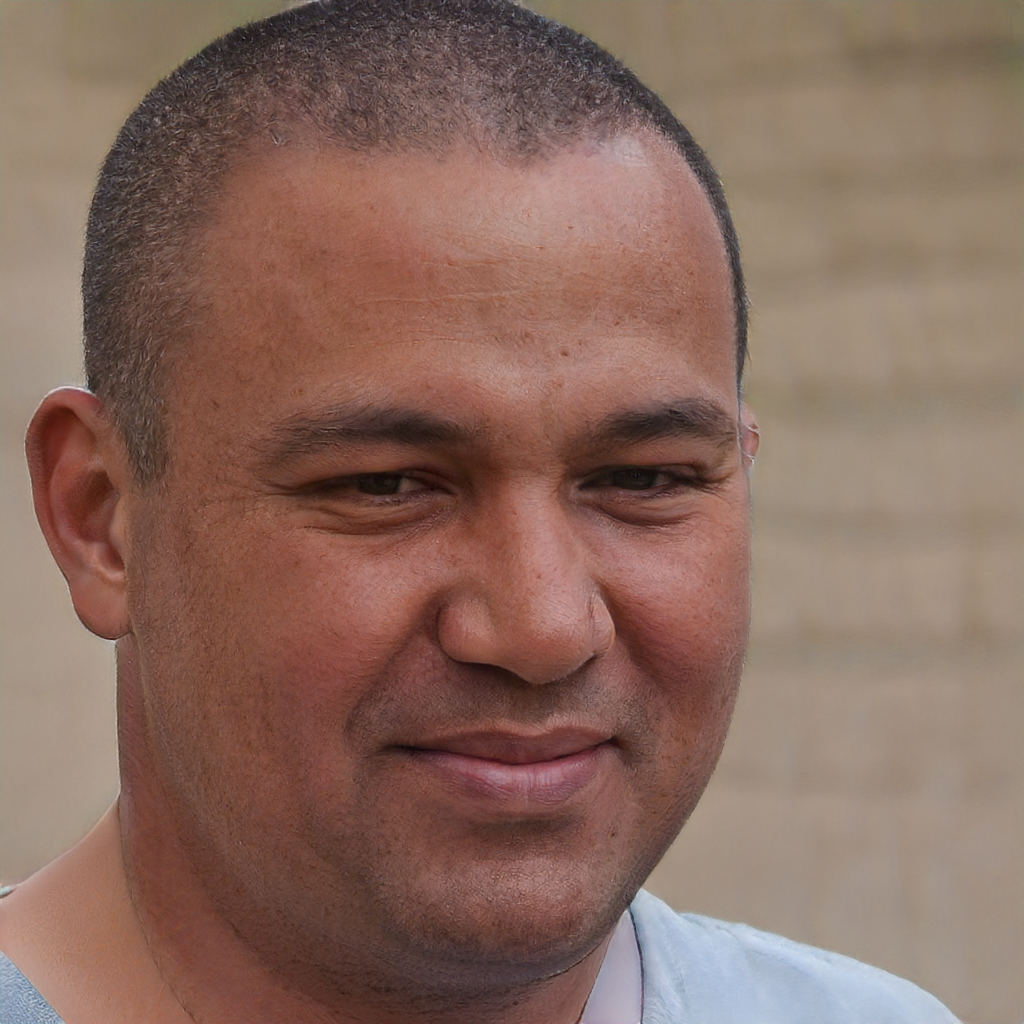}
      \caption{Reconstruction on StyleGAN2}
    \end{minipage}

    \begin{minipage}{0.24\linewidth}
      \centering
      \includegraphics[width=\linewidth]{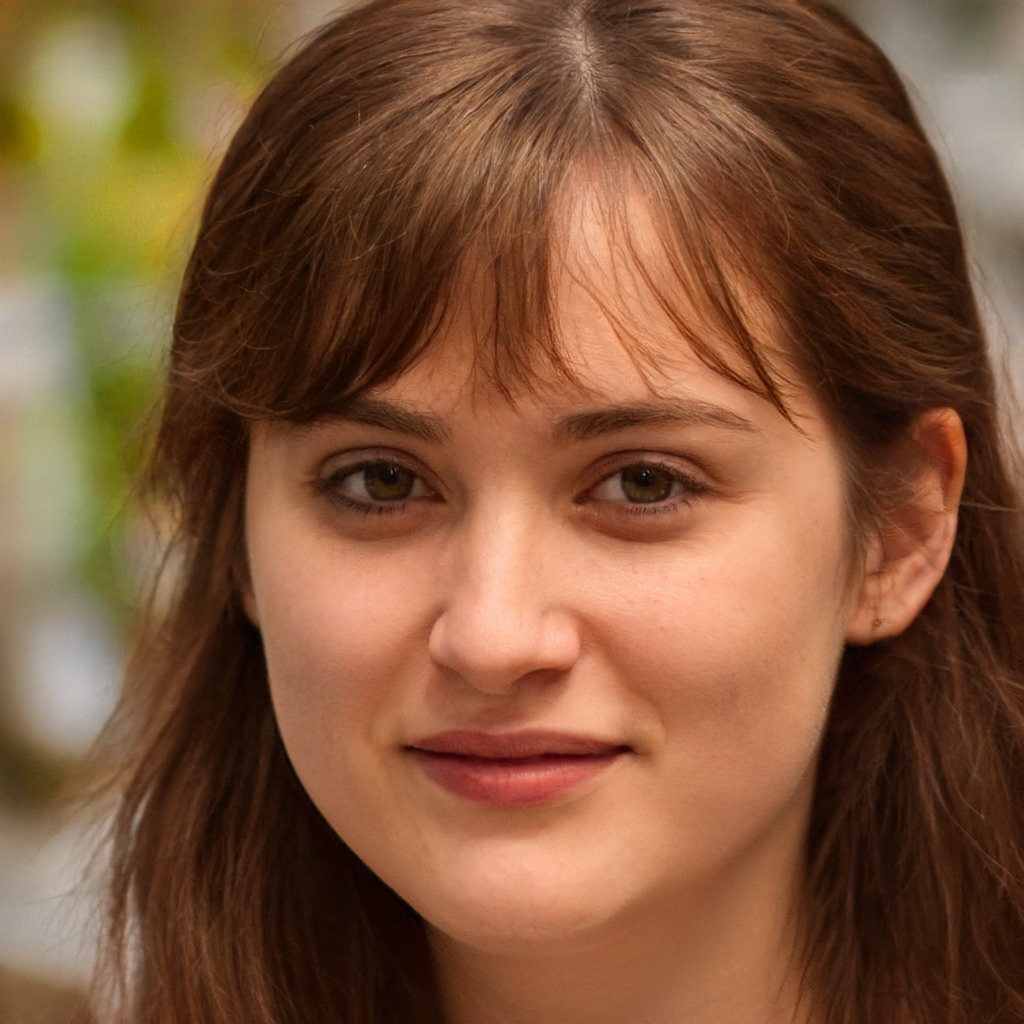}
      \caption{Target (from StyleGAN2)}
    \end{minipage}
    \hfill
    \begin{minipage}{0.24\linewidth}
      \centering
      \includegraphics[width=\linewidth]{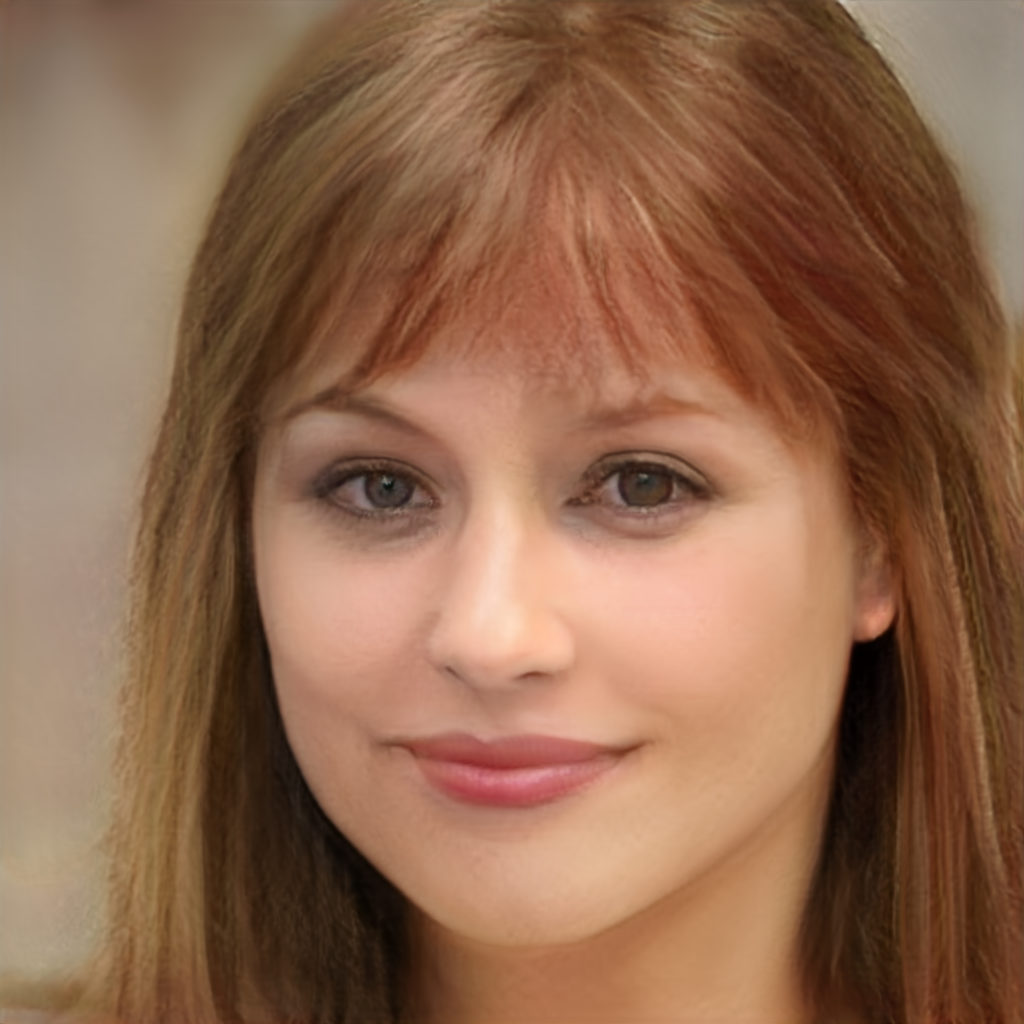}
      \caption{Reconstruction on ProgressiveGAN}
    \end{minipage}
    \hfill
    \begin{minipage}{0.24\linewidth}
      \centering
      \includegraphics[width=\linewidth]{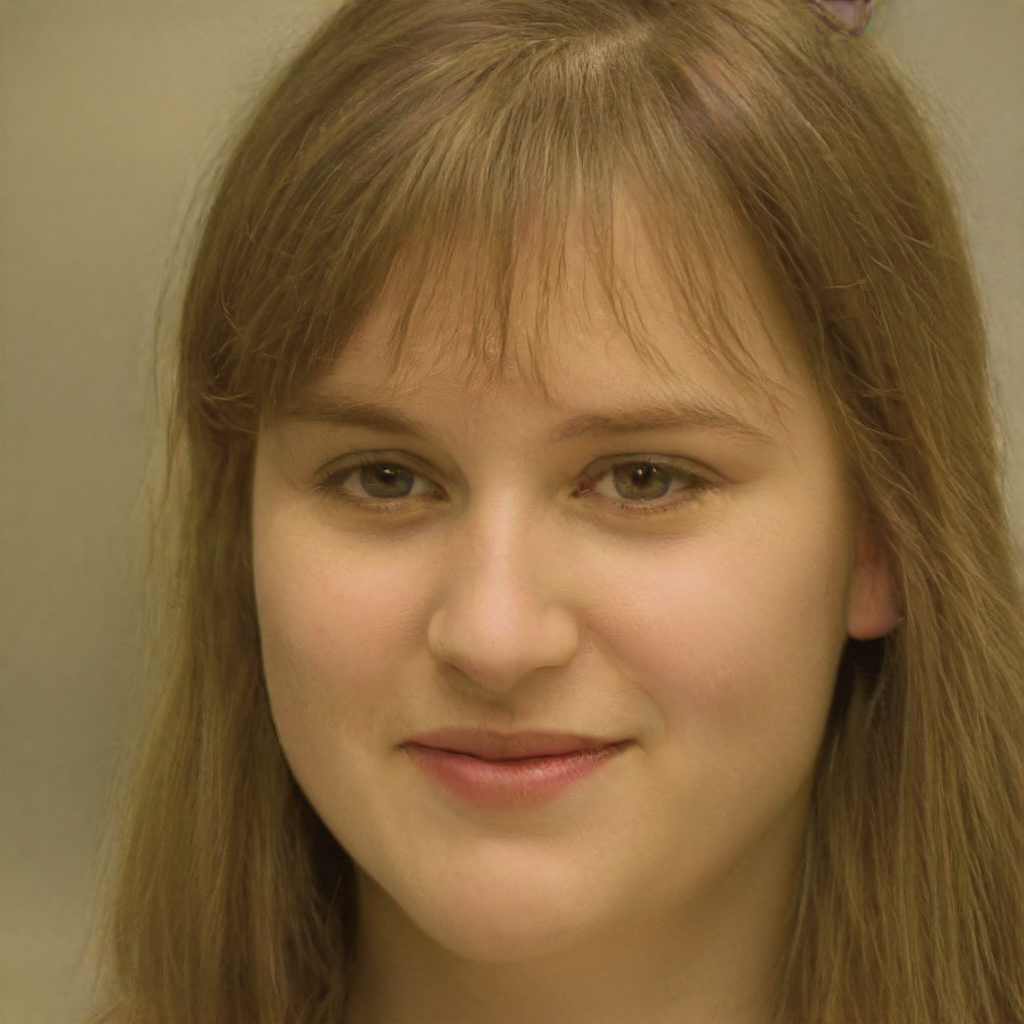}
      \caption{Reconstruction on StyleGAN}
    \end{minipage}
    \hfill
    \begin{minipage}{0.24\linewidth}
      \centering
      \includegraphics[width=\linewidth,cfbox=green 2pt 2pt]{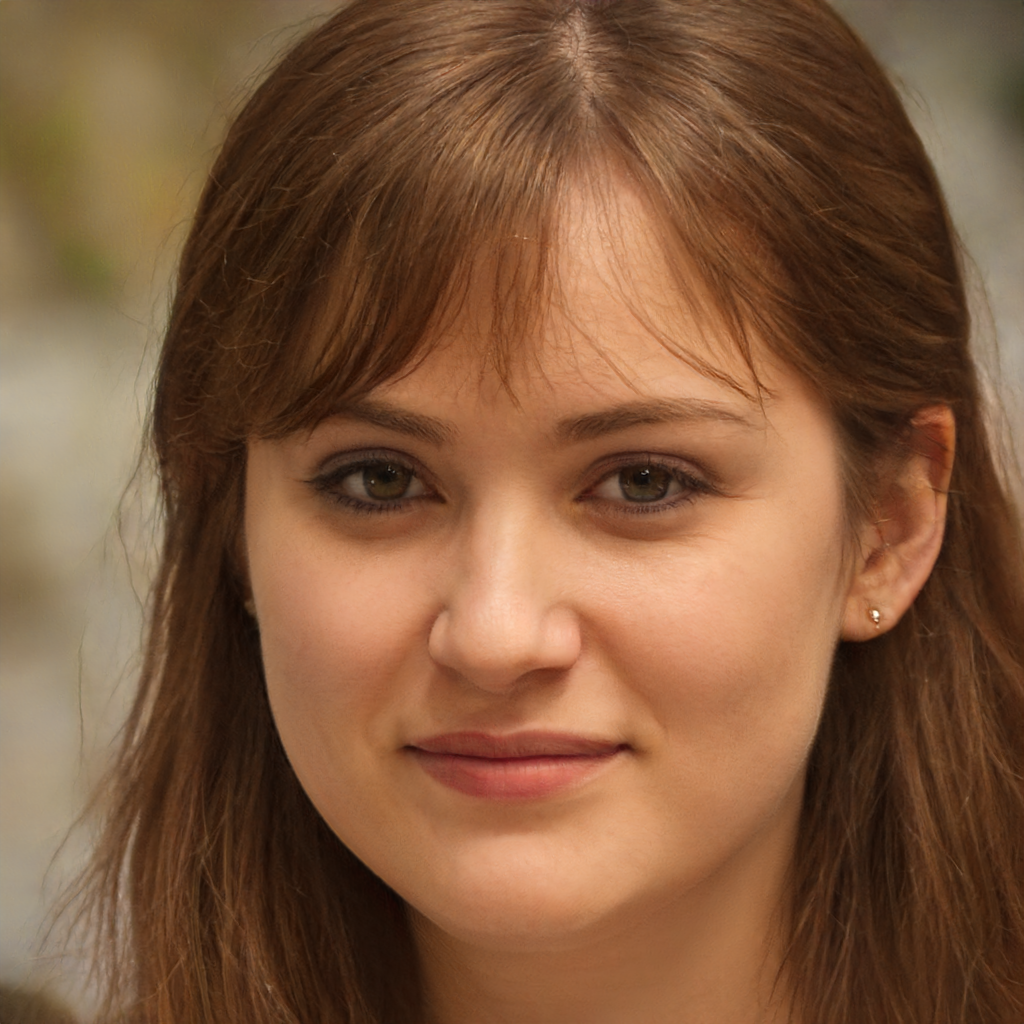}
      \caption{Reconstruction on StyleGAN2}
    \end{minipage}
    
    \begin{minipage}{0.24\linewidth}
      \centering
      \includegraphics[width=\linewidth]{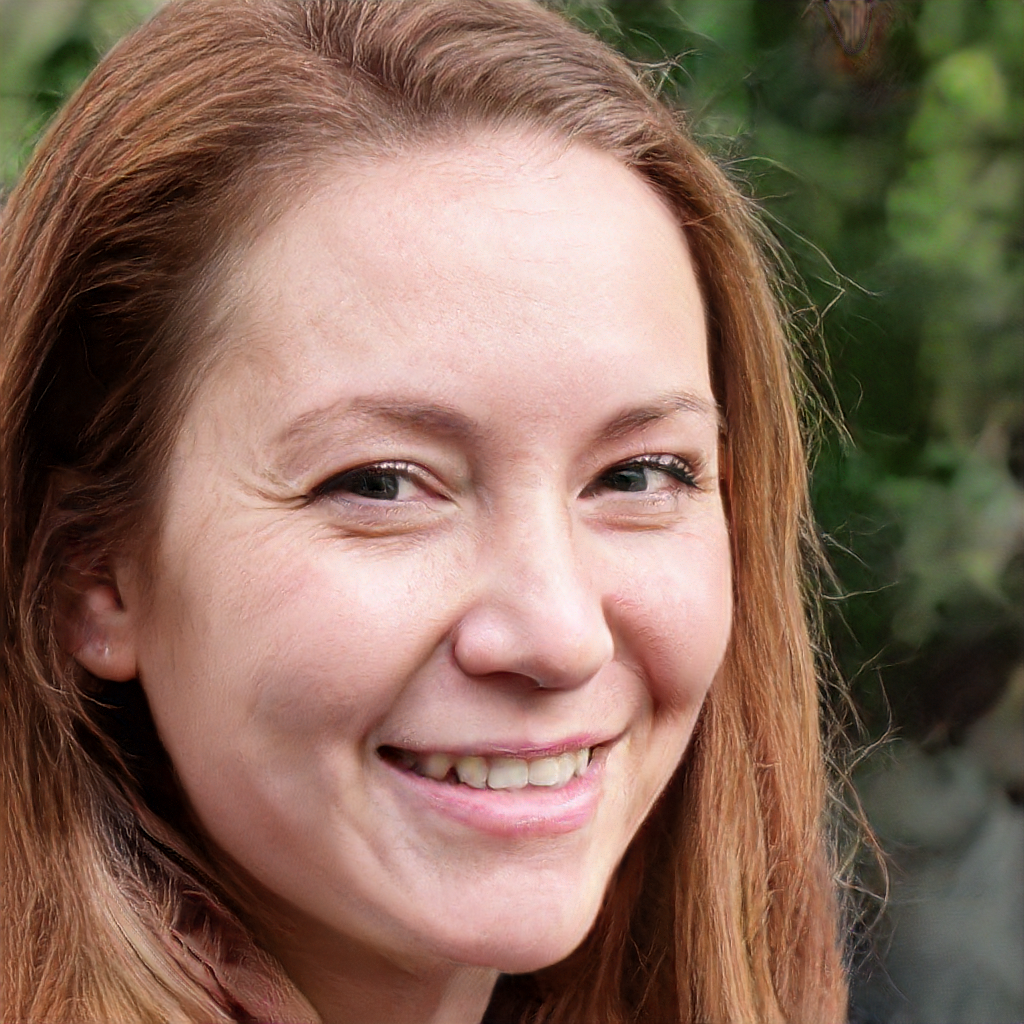}
      \caption{Target (from StyleGAN)}
    \end{minipage}
    \hfill
    \begin{minipage}{0.24\linewidth}
      \centering
      \includegraphics[width=\linewidth,cfbox=green 2pt 2pt]{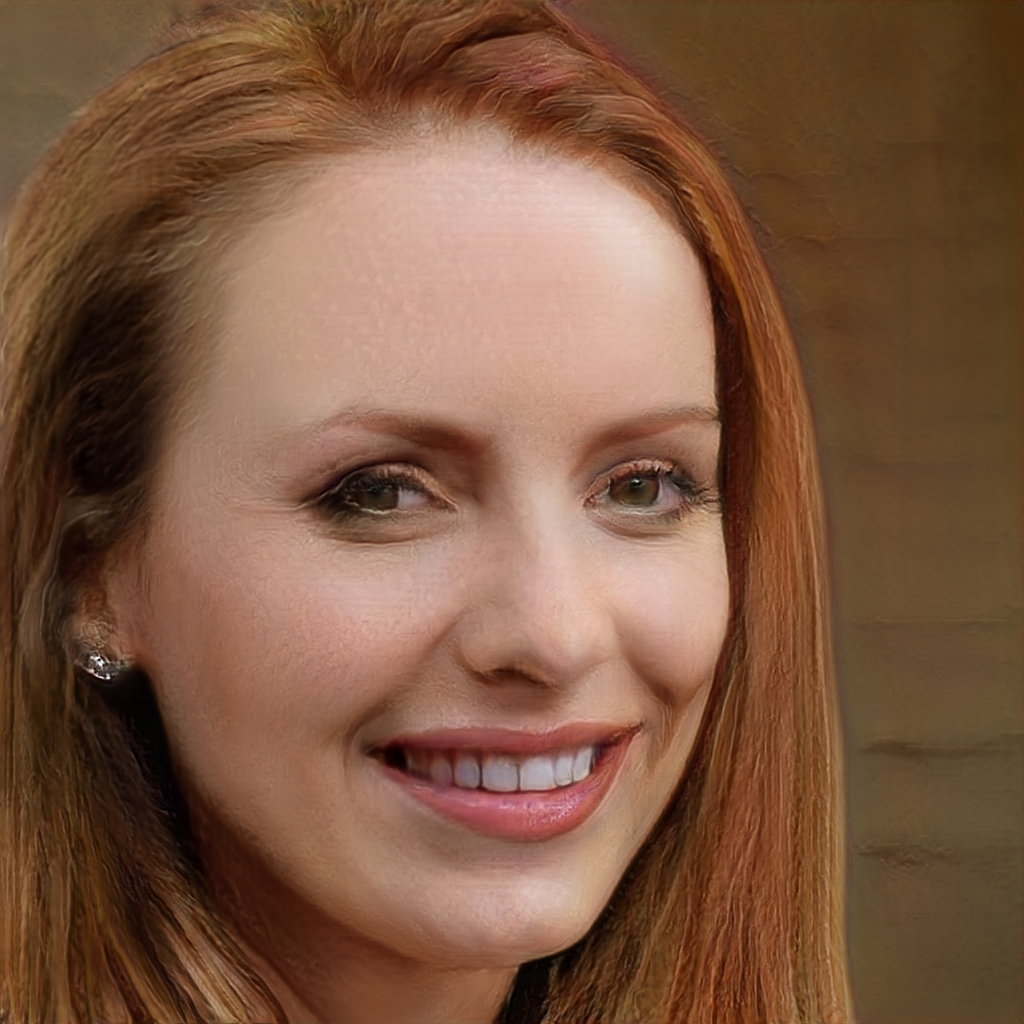}
      \caption{Reconstruction on ProgressiveGAN}
    \end{minipage}
    \hfill
    \begin{minipage}{0.24\linewidth}
        
      \centering
      \includegraphics[width=\linewidth]{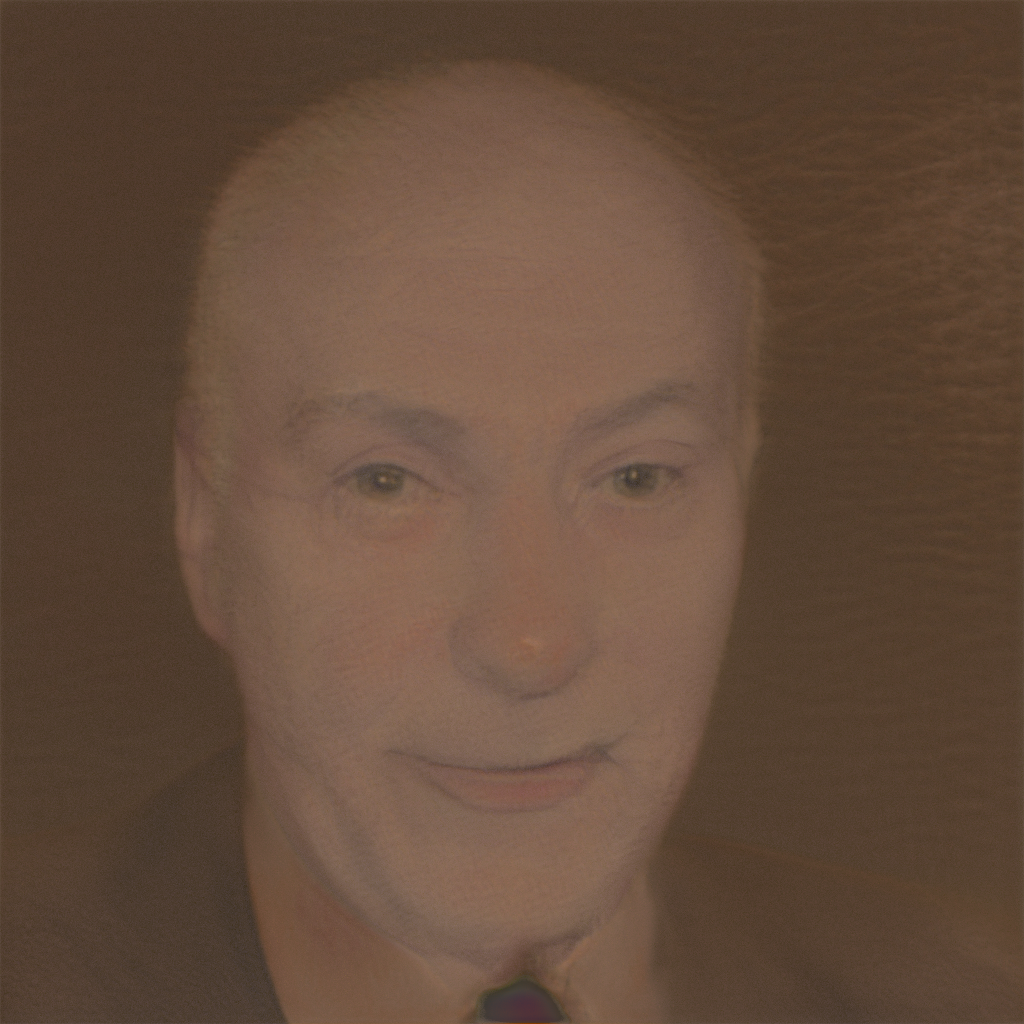}
      \caption{Failed Reconstruction on StyleGAN}
      \label{fig:face_43}
    \end{minipage}
    \hfill
    \begin{minipage}{0.24\linewidth}
      \centering
      \includegraphics[width=\linewidth]{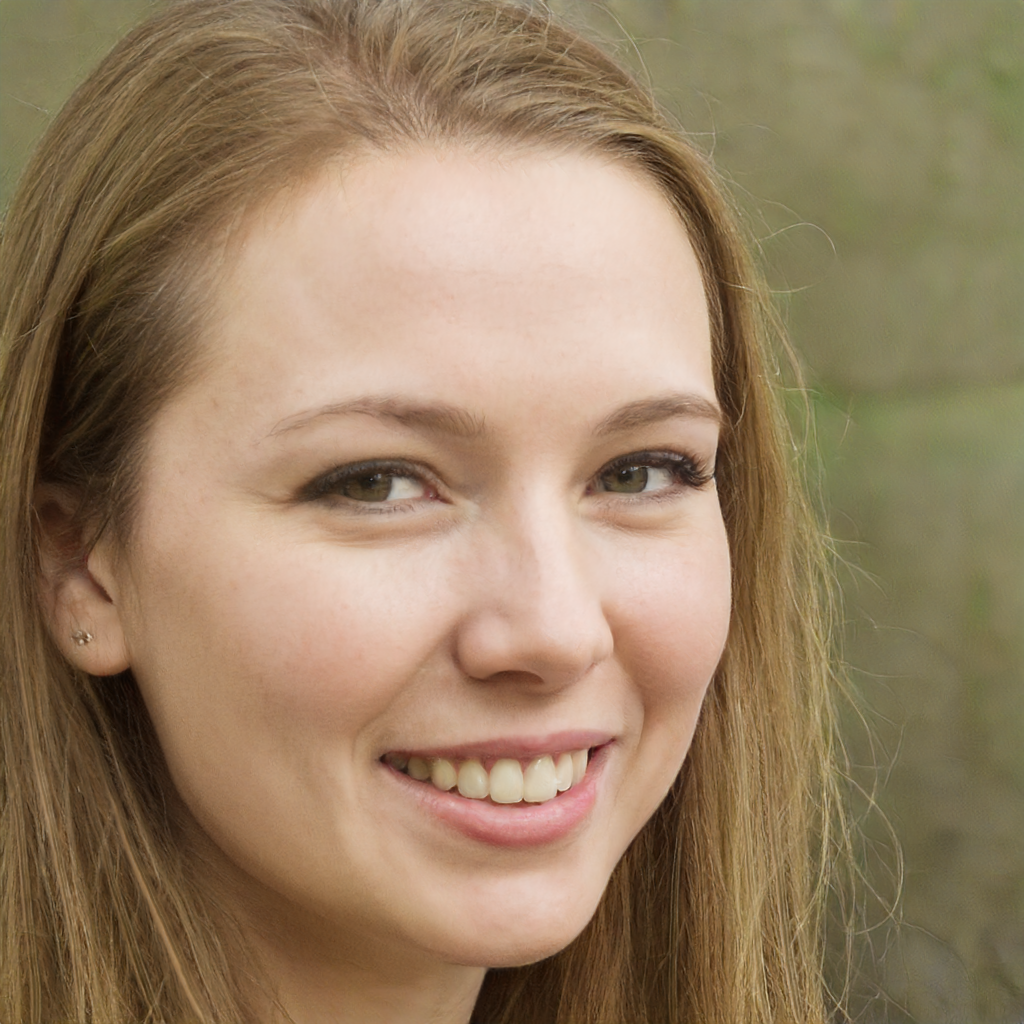}
      \caption{Reconstruction on StyleGAN2}
    \end{minipage}
\end{figure*}

\end{document}